%% file: main.tex
\pdfoutput=1
\documentclass[runningheads]{llncs}

\usepackage{accv}

\usepackage{accvabbrv}

\usepackage{graphicx}
\usepackage{booktabs}
\usepackage{graphicx}
\usepackage{booktabs}
\usepackage{multirow}
\usepackage{multicol}
\usepackage{amsmath}
\usepackage{amssymb}
\usepackage{pifont}
\usepackage{xcolor}
\usepackage{colortbl} 
\usepackage{array}
\usepackage{amsmath}
\usepackage{amssymb}
\usepackage{algpseudocode}
\usepackage{listings}
\usepackage{algorithm}
\usepackage{alltt,xcolor}
\definecolor{lightgray}{gray}{0.9}
\definecolor{lightgreen}{rgb}{0.9, 0.9, 1}
\definecolor{lightred}{rgb}{1, 0.8, 0.8}
\newcommand{\xmark}{{\ding{55}}}%

\usepackage[accsupp]{axessibility}  % Improves PDF readability for those with disabilities.

\usepackage{hyperref}

\usepackage{orcidlink}

\begin{document}

\title{AutoAD-Zero: A Training-Free Framework \\ for Zero-Shot Audio Description} 

\titlerunning{AutoAD-Zero: A Training-Free Framework for Zero-Shot Audio Description}
\author{Junyu Xie\inst{1}\orcidlink{0009-0002-1123-493X} \and
Tengda Han\inst{1}\orcidlink{0000-0002-1874-9664} \and
Max Bain\inst{1}\orcidlink{0000-0002-2345-5441} \and
Arsha Nagrani\inst{1}\orcidlink{0000-0003-2190-9013} \and 
G\"ul Varol\inst{1,2}\orcidlink{0000-0002-8438-6152}\and \\
Weidi Xie\inst{1,3}\orcidlink{0009-0002-8609-6826} \and
Andrew Zisserman\inst{1}\orcidlink{0000-0002-8945-8573} 
}

\authorrunning{J.~Xie et al.}


\institute{Visual Geometry Group, University of Oxford \and
LIGM, \'Ecole des Ponts ParisTech \and
School of Artificial Intelligence, Shanghai Jiao Tong University\\
\url{https://www.robots.ox.ac.uk/vgg/research/autoad-zero/} 
}

\maketitle

\begin{abstract}
Our objective is to generate Audio Descriptions (ADs) for both movies and TV series in a training-free manner. We use the power of off-the-shelf Visual-Language Models (VLMs) and Large Language Models (LLMs), and develop visual and text prompting strategies for this task. Our contributions are three-fold: (i) We demonstrate that a VLM can successfully name and refer to characters if directly prompted with character information through visual indications without requiring any fine-tuning; (ii) A two-stage process is developed to generate ADs, with the first stage asking the VLM to comprehensively describe the video, followed by a second stage utilising a LLM to summarise dense textual information into one succinct AD sentence; (iii) A new dataset for TV audio description is formulated. Our approach, named AutoAD-Zero, demonstrates outstanding performance (even competitive with some models fine-tuned on ground truth ADs) in AD generation for both movies and TV series, achieving state-of-the-art CRITIC scores.
\end{abstract}

\input{sec/01-Introduction}

\input{sec/02-Related_work}
\input{sec/03-Char_prompt}
\input{sec/04-Two-stage_AD_gen}
\input{sec/05-TV_dataset}

\input{sec/06-Experiments}

\input{sec/07-Conclusion}
\input{sec/Acknowledges}
\bibliographystyle{splncs04}
\bibliography{main,vgg_local}

\newpage
\input{sec/99-Appendix}

\end{document}

%% file: sec/01-Introduction.tex
\section{Introduction}
\label{sec:intro}

An Audio Description (AD) soundtrack provides a description of the visual content of a video for the visually impaired.
It covers aspects of the story that cannot be inferred from the audio soundtrack, particularly ``who'' is in the scene and ``what'' they are doing. Furthermore, in order not to overlap with the dialogue, the AD is usually placed in the gaps between characters speaking.
Given the increasing power of Visual-Language Models (VLMs), there has naturally been a growing research interest in automating
the production of ADs for movies.
Most recent approaches have either fine-tuned components of an
open-source pre-trained large-scale VLM, such as the visual-text bridge, for the
AD task~\cite{Han23,Han23a,Han24,wang2024contextual,srivastava2023you,raajesh2024micap}, or have used very powerful propriety models such as GPT-4/GPT-4v
in a zero-shot way, without fine-tuning~\cite{Zhang_2024_CVPR,Lin2023mmvid,chu2024llmadlargelanguagemodel}. 
Both approaches have limitations: there is
insufficient data to fine-tune these large models for the AD task -- to enable the model to distill out the essential information
for AD and ignore the rest, and
current zero-shot techniques fail due to a lack of knowledge of who the characters are, a core requirement for AD.

\begin{figure}[t]
    \centering
    \includegraphics[width=0.95\linewidth]{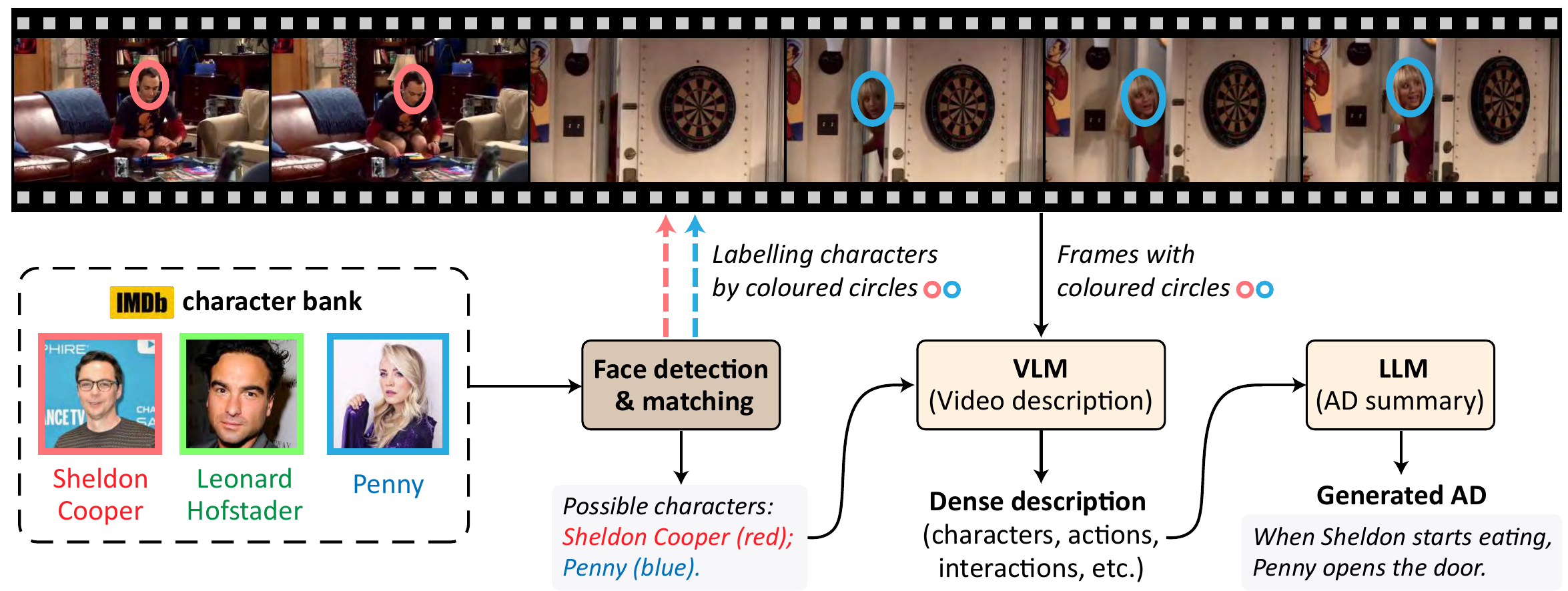}
    \vspace{-0.15cm}
    \caption{\textbf{A training-free framework for zero-shot AD generation.} AutoAD-Zero features a two-stage process, where a VLM initially generates a comprehensive video description from multiple aspects, followed by an LLM-based AD summary in the second stage. To incorporate character information into this framework, character faces in the input video are matched with those in an external character bank and labelled with coloured circles. The corresponding character names and colour codes are then provided as text prompts to the VLM.}
    \label{fig:teaser}
\end{figure}

In this paper, we show that it is possible to adopt pre-trained models with novel visual and text prompting. Somewhat surprisingly, this zero-shot, training-free method reaches competitive performance with all previous AD approaches, and even gets state-of-the-art in some metrics. 
To achieve this we 
explore two ideas, as demonstrated in~\cref{fig:teaser}: 
first, use a {\em visual prompt} on the
video frames so that we can refer to characters by name when text
prompting the VLM. We achieve this by ``circling'' the characters in
the frame and color coding for each identity.  This ``circle'' prompting~\cite{Shtedritski23}
enables the VLM both to identify the characters in the clip, and also
to refer to them by name when describing their actions, interactions,
and locations. The second idea is {\em not} to predict the AD
directly, but instead to take advantage of the strength of Large Language Models (LLMs) at providing summaries. We introduce a two-stage approach to generating the AD: in the first stage, a detailed description of ``who is doing what'' is generated by prompting the VLM with questions about the
visually tagged characters; in the second stage, an LLM is used to summarise the detailed description into the style and desired length for the AD.
This two-stage approach has multiple advantages -- 
it forces the AD to be more visually grounded;
it enables the information in the AD to be prioritised for the temporal
interval available; 
it enables the style of AD to be tailored for particular groups, such as being age appropriate~\cite{salewski2024context} or a simpler form of the language (for non-native speakers).

The great advantage of our training-free approach, is that the method can be directly 
plugged into new VLMs as they are released (which is very often at the moment) to benefit from improvements in their descriptive power. It can also be plugged into propriety models because no training is required.

We also investigate generating AD automatically for TV series, 
such as \textit{``Friends''} and \textit{``The Big Bang Theory''}. This domain differs from movies in a number of aspects: a TV series can often have a cast of characters that is persistent over multiple episodes; 
and due to shorter episodes, the series often have denser dialogues than films. In older series, the cinematography can also differ, with more close-ups and mid-shots in TV than in movies, whereas there may be more long shots in movies.
To this end, we introduce a new dataset of TV series (TV-AD) with ground truth aligned AD annotations.

In summary, we make the following contributions: (i) We devise a visual-textual prompting mechanism to incorporate character information into the VLM in a training-free manner; (ii) We propose a two-stage AD generation process, where a comprehensive character-aware video description is first generated by a VLM, followed by an LLM-based summary stage to obtain the final AD; (iii) We formulate a new TV-AD dataset and investigate the nature of TV ADs in comparison to their movie counterparts; (iv) The proposed AutoAD-Zero model is evaluated on multiple movie/TV AD datasets, including MAD-Eval~\cite{Han23}, CMD-AD~\cite{Han24}, and TV-AD, significantly outperforming existing training-free AD generation methods.  When compared to models explicitly fine-tuned on ground truth AD annotations, AutoAD-Zero demonstrates competitive performance and achieves state-of-the-art results on the CRITIC scores.

%% file: sec/02-Related_work.tex
\section{Related Work}
\label{sec:rel_work}
\noindent \textbf{Previous AD and Video Captioning Models.}
With the power of large vision and language models, 
rapid progress has been made in video captioning.
The dense video captioning task focuses on short video clips~\cite{lin2022swinbert,luo2020univilm,seo2022end},
and other variants summarise longer video segments that span a few minutes~\cite{lu2013story,zhang2016summarization}.
The task of Audio Description (AD) generation is closer to dense video captioning with some differences,
where the model aims to describe visual elements in the movie or TV densely over time, in a story-telling manner.
Unlike video captioning, AD generation requires a highly summarised output with strict formatting, coherent context~\cite{Han23,wang2024contextual,ye-etal-2024-mmad-multi}, and character information~\cite{Han23a,raajesh2024micap}. 
Notably in~\cite{Han23a}, a separate character recognition module is trained to feed character identities to the VLM, which is shown to be critical compared with an alternate method that applies GPT-4 model for AD generation~\cite{Zhang_2024_CVPR}.
Character recognition in movies or TV typically can be achieved by face matching with portrait exemplars~\cite{huang2018person} or joint audio-visual matching that includes voice~\cite{korbar2024look}.
The emotion of the characters is also explored for movie descriptions~\cite{srivastava2023you}.

On the data acquisition side,
due to the copyright constraints of movies and TV series,
early works share short video clips~\cite{rohrbach2015lsmdc} or pre-computed visual features~\cite{soldan2022mad}, limiting the scale and power of AD generation models.
\cite{Han24} proposes to align YouTube videos from \cite{Bain20} with AD annotations to acquire longer movie clips with pixels. A recent work~\cite{ghermi2024short} collects datasets with amateur-made short movies (13 mins on average) that are publicly available and easier to share.

\vspace{3pt}
\noindent \textbf{Prompting for LLMs and VLMs.} 
Prompting refers to designing instructions that guide the pre-trained models to generate desired outputs.
One stream of works learns vector prompts with targeted training data, on both LLMs~\cite{li2021prefix,lester2021prompt} and VLMs~\cite{zhou2022coop,zhou2022cocoop,ju2022prompting}.
Another stream designs text-only prompts which is more compatible with black-box pre-trained models prohibiting any model modifications.
For example, it is found that adding a text prompt ``let's think step by step''~improves the performance of GPT3 on various tasks \cite{kojima2022stepbystep}.
Using text prompts with chain-of-thought examples can activate the reasoning in LLM's answers~\cite{wei2022chain}.

Our work is closer to ``visual prompting'', that provides supplementary \emph{visual} information to the VLMs without any fine-tuning~\cite{jia2022vpt,bar2022visual}. In~\cite{Shtedritski23}, the authors explore visual prompting for large-scale VLMs like CLIP, discovering that drawing a red circle around an object can effectively direct the model's attention, 
leading to strong performance in zero-shot referring expressions comprehension, and keypoint localization tasks.
In~\cite{cai2023vipllava}, the authors introduce a multimodal model that decodes arbitrary visual prompts, i.e.\ visual markers overlayed on the RGB image, allowing intuitive interaction with images through natural cues such as ``red bounding box'' or ``pointed arrow''. 
In~\cite{yang2023set}, the authors introduce Set-of-Mark (SoM), a visual prompting method that enhances the visual grounding capabilities of large multimodal models like GPT-4v. By using interactive segmentation models to partition images and overlaying them with marks, SoM enables models to answer questions requiring visual grounding. 

\vspace{3pt}
\noindent \textbf{Training-Free VLM Adaptation.} 
With the strong performance and huge training cost of large-scale pretrained VLMs,
many works attempt to adapt VLMs to specific tasks in a training-free manner.
Yan~\etal~\cite{yang2023dawn} showcases GPT-4v on a wide range of multi-modal applications, 
such as recognizing celebrities and famous landmarks, and understanding code snippets from screenshots.
State-of-the-art VLMs are also adapted to various medical applications~\cite{wu2023medical,yan2023medical}, but these works find the VLMs still lack expertise in medical knowledge.

Our work is closely related to MM-Narrator~\cite{Zhang_2024_CVPR} and LLM-AD~\cite{chu2024llmadlargelanguagemodel}, where MM-Narrator leverages GPT-4/GPT-4v 
with multi-modal in-context learning to generate the audio descriptions for long movies, in an auto-regressive manner with memories. A concurrent work, LLM-AD, applies visual prompting to GPT-4v by overlaying character names onto movie frames. In contrast, we introduce a visual-textual prompting approach that provides character names as a part of the text prompts with colour-coded circles as fine-grained visual indicators, leading to superior performance. 

%% file: sec/03-Char_prompt.tex
\section{Prompting VLM with Character Information}
\label{sec:char_prompt}
\begin{figure}[t!]
    \centering
    \includegraphics[width=0.95\linewidth]{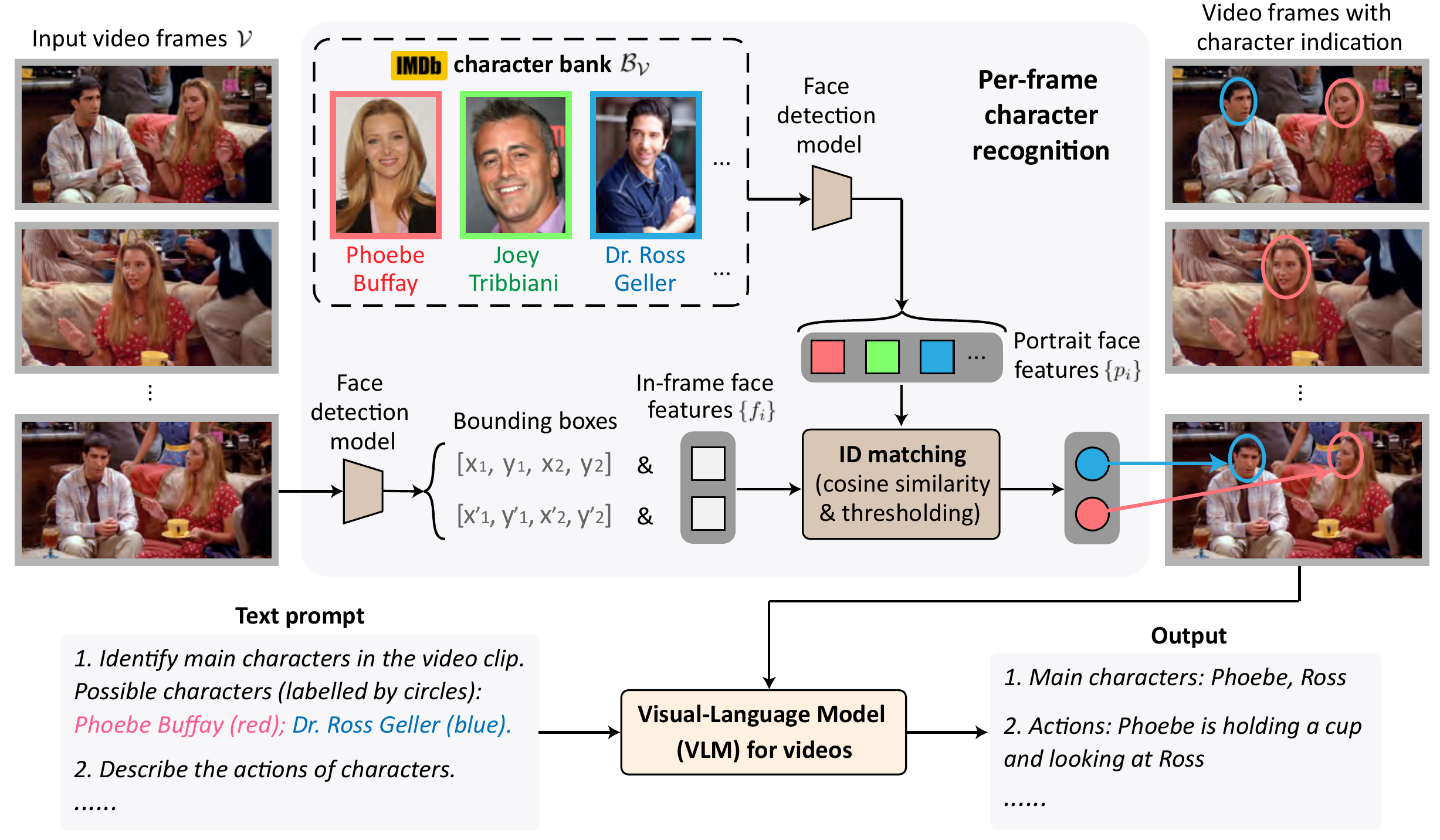}
    \vspace{-0.15cm}
    \caption{\textbf{Character recognition and VLM prompting}. An off-the-shelf face detection model is employed to obtain bounding boxes and face features in video frames. These ``in-frame face features'' are then matched with ``portrait face features'' extracted from character profile images, which determines the identities in the video. To prompt the VLM with character information, character faces are labelled by coloured circles, with corresponding names and colour codes provided in the text prompt.}
    \label{fig:char_prompt}
\end{figure}
To incorporate character information into ADs, prior works~\cite{Han23a, Han24, wang2024contextual} explicitly tune the VLM to adapt to character information. We instead explore a training-free route that involves two steps, as detailed in~\cref{fig:char_prompt}: (i) character recognition based on an off-the-shelf face detection model, then (ii) prompt VLM with visual character indications.

\subsection{Character Recognition by Face Detection} 
\label{subsec:face_det}
Given a video clip $\mathcal{V}$ from a movie or a TV series episode, we query the corresponding cast list from the database. The video-specific character bank can be then obtained as $\mathcal{B}_{\mathcal{V}} = \{[\texttt{char}_i, \mathcal{P}_{i}]\}_{i=1}^{C}$ where there are $C$ characters, $\texttt{char}_i$ indicates the character name, and $\mathcal{P}_{i}$ denotes the profile portrait image of the corresponding actor. An exemplary character bank shown in~\cref{fig:char_prompt} is:
\begin{center}
    \{[\textit{``Phoebe Buffay''}, $\mathcal{P}_{1}$], [\textit{``Joey Tribbiani''}, $\mathcal{P}_{2}$], [\textit{``Dr. Ross Geller''}, $\mathcal{P}_{3}$], ...\}
\end{center}
We then formulate a pipeline to detect faces in the video frame and match them with those in the character bank. This process is \emph{independently} conducted for each frame.

\vspace{3pt}
\noindent \textbf{Face Feature Extraction.} 
Given a video frame $\mathcal{I}$ in the clip $\mathcal{V}$, an off-the-shelf face detection model $\Phi$ is employed to identify faces: 
\begin{equation}
    \{\texttt{Bbox}_j, f_j\}_{j=1}^{N} = \Phi(\mathcal{I})
\end{equation}
where $N$ faces are detected in a frame, with the feature embedding of each face $f_j$ extracted, together with a predicted bounding box coordinate $\texttt{Bbox}_j$. We refer to the set of face embeddings as \emph{in-frame face features} $\{f_i\}_{i=1}^{N}$. 

Similarly, a face embedding is extracted from each portrait image $\mathcal{P}_i$ in the character bank, together forming a set of \emph{portrait face features} $\{p_i\}_{i=1}^{C}$.

\vspace{3pt}
\noindent \textbf{Face ID Matching.} 
To match the faces within video frames with those in portrait images, we evaluate the cosine similarities $\mathcal{A}_{ij}$ between two sets of features and identify the best-matched portrait face for each in-frame face feature. Formally,
\begin{equation}
    \texttt{ID}_j = \text{max}_i(\mathcal{A}_{ij}), \;\;\;\text{where} \;\;\mathcal{A}_{ij} =\frac{p_i \cdot f_j}{|p_i||f_j|}
\end{equation}
where $\texttt{ID}_j$ denotes the predicted face ID in the character bank. As a result, the predicted bounding boxes within the video frame could be matched with a character name, forming a triplet set $\{[\texttt{char}_{\texttt{ID}_j}, \texttt{Bbox}_j, \mathcal{A}_{\texttt{ID}_jj}]\}_{j=1}^{N}$, where $\mathcal{A}_{\texttt{ID}_jj}$ is the matching score. 

We filter out triplets based on the matching scores and the following rules:
\begin{itemize}
    \item If $\mathcal{A}_{\texttt{ID}_jj} < \epsilon$, where $\epsilon$ denotes a threshold, this indicates that the face matching is not confident, possibly owing to poor video quality, motion blur, non-important characters' faces, \emph{etc.} We therefore remove those faces (triplets) from the set.
    \item If the predicted IDs for multiple faces are the same within a frame, we keep the one with the highest matching score $\mathcal{A}_{\texttt{ID}_jj}$.
\end{itemize}

\subsection{Visual Prompting VLM by Circling Character Faces} 
\label{subsec:vlm_char_prompt}
In the training-free design, one key challenge lies on the strategy to prompt VLM for understanding the association between visible characters and their names. Recent work~\cite{Shtedritski23} has revealed that drawing a red circle (ellipse) could effectively direct the CLIP's attention to a local region while maintaining global information understanding. Inspired by this, and given the fact that CLIP is popularly adopted as the visual encoder for recent VLMs, we overlay coloured circles around character faces in the frame as direct visual indications.

As shown in~\cref{fig:char_prompt}, we assign a unique colour code for each character, and transform the corresponding bounding box coordinates into circles that are overlaid on the raw video frames. These frames with coloured circles are then directly taken as input to a VLM, while the character names and corresponding colours are listed in the input text prompts, e.g.\ \textit{``Phoebe Buffay (red)''}. 
By leveraging coloured circles as a bridge, it can be observed that the VLM is capable of associating character information between text and video and using it to accomplish other tasks, such as action description.

%% file: sec/04-Two-stage_AD_gen.tex
\section{Two-Stage Training-Free AD Generation}
In this section, we introduce a training-free method to generate AD sentences from video frame inputs. This approach consists of two stages: (i) comprehensive character-aware video description by a VLM, and (ii) AD summary by an LLM based on the outputs of the first stage.

\label{sec:ad_gen}
\begin{figure}[t!]
    \centering
    \includegraphics[width=0.95\linewidth]{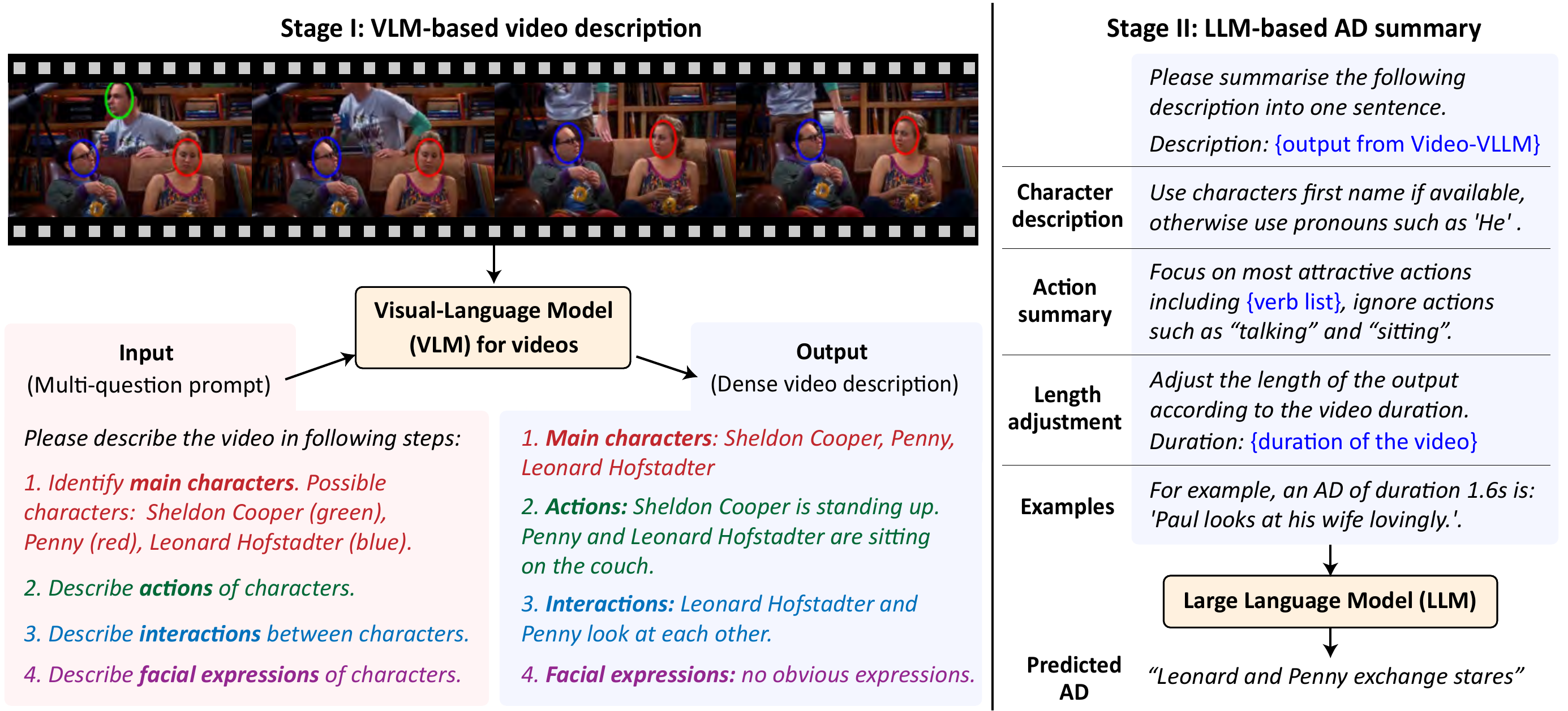}
    \vspace{-0.15cm}
    \caption{\textbf{Two-stage training-free AD generation}. The first stage adopts a VLM to produce a comprehensive video description, covering aspects including main characters, actions, interactions, and facial expressions. The second stage uses an LLM to summarise the video into a single AD sentence, extracting the most relevant character and action information, and adjusting the content and style according to specific rules.}
    \label{fig:ad_gen}
\end{figure}

\subsection{Stage I: VLM-Based Video Description}
The VLM processes video frames along with a list of text-form questions (e.g.\ who are the characters? and what are they doing?) to generate a comprehensive video description. Compared to directly outputting a one-sentence AD, this process adopts the Chain-of-Thought (CoT) approach, guiding the model to systemically understand the video.

\cref{fig:ad_gen} (left) provides a simplified version of the input text prompt, where we begin by asking the model to \textbf{identify the main characters} (``who'') in the video (point 1). Following the process described in~\cref{subsec:vlm_char_prompt}, we label individual characters with colored circles and refer to them in the text by specifying the colors, such as \textit{``Sheldon Cooper (red)''}. We found that by first asking the model for character names, it effectively remembers and naturally uses these names for describing actions.

We then proceed with questions about ``what'', 
that consists of three major directions, asking the model to:
\begin{itemize}
    \setlength\itemsep{0.3em}
    \item Describe the \textbf{actions} of characters (point 2). This includes characters' states (e.g.\ sitting or standing), gestures, movements (e.g.\ walking), and interactions with objects (e.g.\ pouring a drink).
    \item Describe the \textbf{interactions} between characters (point 3). Despite potentially overlapping with point 2, this point emphasises the importance of focusing globally on multiple characters and understanding their interactions, including physical contacts (e.g., hugging) and distant communications (e.g.\ looking at others).
    \item Describe the \textbf{facial expressions} of characters (point 4). This instruction encourages the model to focus on facial details that may convey the characters' emotions. Furthermore, directing the model to zoom in on facial regions enhances the accuracy of character recognition.
\end{itemize}

Additional directions on describing the environment and character appearances are also investigated, but no significant improvements are observed (see ablations in~\cref{subsec:result_ad_pred}).

These points together form a multi-question prompt that is fed into VLM along with the raw video frames. We ensure that the model answers are in the same order as the questions (the template is shown in~\cref{fig:ad_gen}), resulting in a dense video description. One can also ask the VLM model to summarise the description into a one-sentence AD at the end. However, we found it challenging to enforce its adherence to the AD format while simultaneously generating high-quality descriptions.

\subsection{Stage II: LLM-Based AD Summary}
As demonstrated in~\cref{fig:ad_gen} (right), the second stage aims to summarise the video description obtained from the first stage into a single AD sentence, while adjusting the outputs according to specific rules. Note that, we use an LLM for this task that takes text-form inputs only (no video inputs).

Specifically, the text prompt starts with \textit{``Please summarise the following description into one sentence.''}, followed by the provision of first-stage outputs. The remaining instructions contain four main components:
\begin{itemize}
    \setlength\itemsep{0.3em}
    \item \textbf{Character description}, which formalises the use of character names (i.e.\ only the first name) and pronouns, rather than descriptions such as ``a man''.
    \item \textbf{Action summary}, which asks the LLM to extract the most important characters and their actions. To guide this process, we extract top-$k$ ($k=15$) most frequent verbs in \emph{training splits} of AD datasets to form a verb list. Additionally, actions related to static states (e.g.\ sitting) and talking-related actions are considered non-informative for ADs, as they can be inferred from previous contexts or dialogues.
    \item \textbf{Length adjustment}, which provides the duration of the time windows for ADs as hints that indicate how much detail should be included in the output.
    \item \textbf{Examples}, which offer references for the LLM regarding AD styles and lengths.
\end{itemize}

After combining all these factors into the text prompt, the LLM outputs a single sentence as the summarised AD.
 

%% file: sec/05-TV_dataset.tex
\section{TV-AD Dataset}
\label{sec:tvad_dataset}
In this section, we introduce a new dataset, namely TV-AD, that contains ground truth audio descriptions for TV series. \cref{subsec:dataset_detail} specifies the dataset details and provides a comparison with a movie AD dataset. \cref{subsec:dataset_formulation} elaborates on how the dataset is formulated by aligning AD annotations with TV episodes.
\subsection{Dataset Details}
\label{subsec:dataset_detail}
\cref{tab:dataset} provides statistics of the TV-AD dataset, which features episodes across multiple TV series including \textit{``The Big Bang Theory'', ``Friends'', ``Frasier'', ``Seinfeld''}, etc. (The full list is available in~\Cref{supsec:tvad}.)
We further divide the dataset into training (TV-AD-Train, $\sim31$k ADs) and evaluation splits (TV-AD-Eval, $\sim3$k ADs), ensuring that the TV series do not overlap between the two splits. The evaluation split contains AD annotations for TV videos that are publicly available (from TVQA~\cite{lei2018tvqa}) and will be publicly released.

{\vspace{6pt}
\noindent\begin{minipage}[t!]{0.48\textwidth}%
\centering
\setlength\tabcolsep{2pt}
\resizebox{\textwidth}{!}{
\begin{tabular}{cccccc}  
\toprule
\multirow{2}{*}{Split}  & \multirow{2}{*}{\;\shortstack{TV\\series}\;}  &  \multirow{2}{*}{Seasons} &  \multirow{2}{*}{Episodes}  & \multirow{2}{*}{ADs} & \multirow{2}{*}{\;\shortstack{Total\\duration}\;}\\
 & &   &   &  & \\
\midrule
Train & $11$ & $18$ & $326$  & $31030$ & $214$h  \\
Eval & $2$ & $5$  & $100$  & $2983$ &  $37$h \\
\midrule
Total & $13$ & $23$  & $426$  & $34013$ &  $251$h \\
\bottomrule
\end{tabular}}
\captionof{table}{\textbf{Statistics of the TV-AD dataset.} Train and Eval splits consist of different TV series. 
}
\label{tab:dataset}
\end{minipage}
\hspace{0.2cm}
\noindent\begin{minipage}[t!]{0.48\textwidth}%
\includegraphics[width=0.95\linewidth]{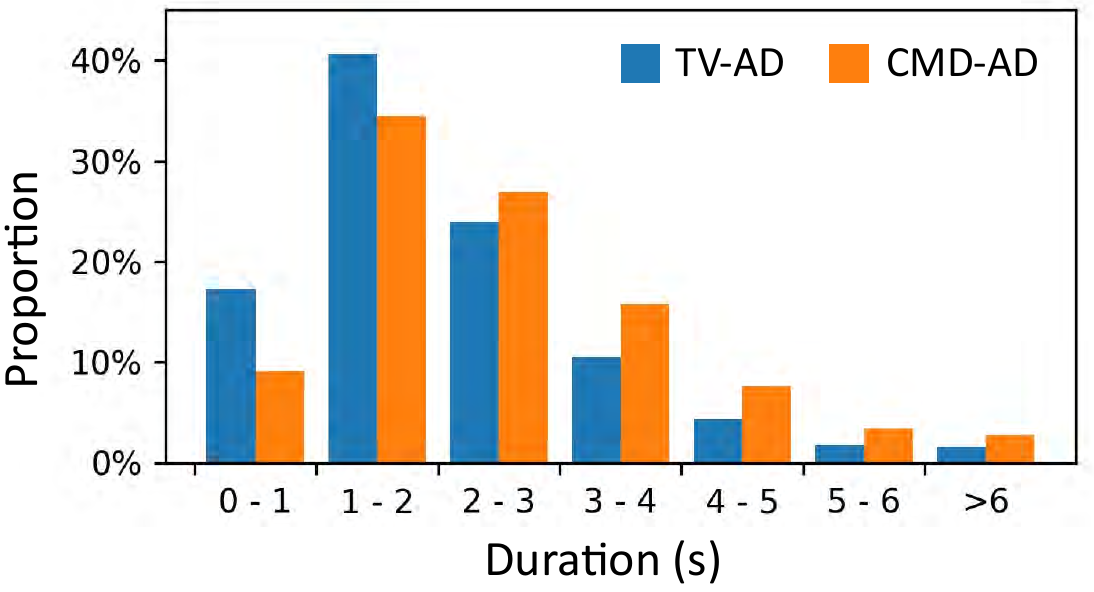}
\vspace{-0.3cm}
\captionof{figure}{\textbf{Comparison of AD duration} between TV-AD and CMD-AD.}
\label{fig:dataset_comparison}
\end{minipage}
\vspace{3pt}
}

\vspace{3pt}
\noindent \textbf{Comparison with CMD-AD.} CMD-AD~\cite{Han24} (where ``CMD'' stands for Condensed Movie Dataset~\cite{Bain20}) is a movie AD dataset that contains $\sim100$k ADs from $\sim1.4$k movies. The duration of ADs between TV-AD and CMD-AD are compared in~\cref{fig:dataset_comparison}. Movie ADs (mean duration = $2.51$s) are shown to be generally longer, while TV ADs (mean duration = $2.07$s) consist of a large proportion within $1-2$s. This could be due to the shorter time of each TV episode compared to movies, as well as fast-paced plots and dialogues in TV series, resulting in more compressed audio descriptions.

\subsection{Dataset Formulation: Aligning ADs with TV Series}
\label{subsec:dataset_formulation}
\begin{figure}[t!]
    \centering
    \includegraphics[width=0.95\linewidth]{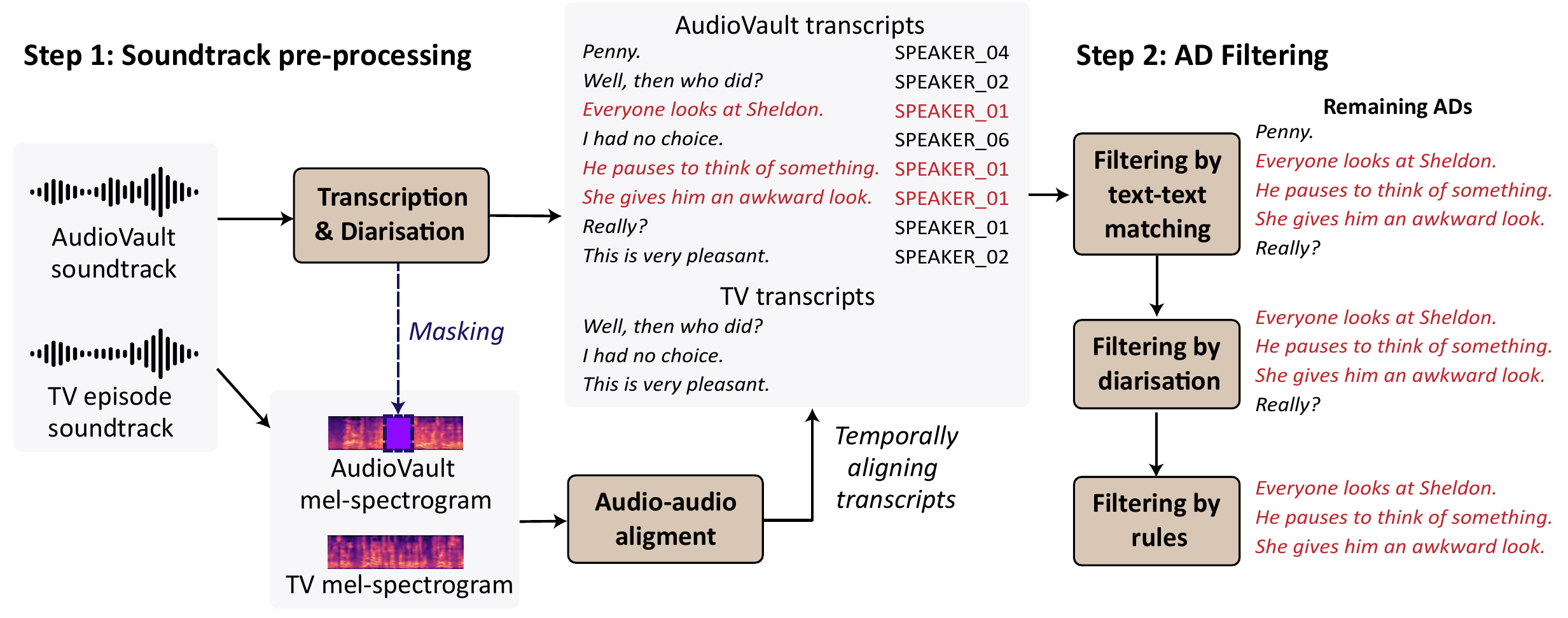}
    \vspace{-0.15cm}
    \caption{\textbf{Dataset formulation}. The aim is to convert AD annotations from AudioVault into text form and align them with TV episodes. The main pipeline consists of two steps: (i) The soundtrack pre-processing step, which aligns the AudioVault and TV timestamps via audio-audio matching and transcribes both sound sources into transcripts; (ii) The AD filtering step, which retrieves text-form ADs from the transcripts and performs further cleaning.}
    \label{fig:ad_align}
\end{figure}

The AudioVault website\footnote{\url{https://audiovault.net/}} (as also used by \cite{soldan2022mad,Han23,Han24}) provides human-annotated audio descriptions for thousands of movies and TV series in the form of audio files, with the spoken ADs merged with the original TV soundtracks. To formulate the TV-AD dataset, we (i) temporally align the AudioVault soundtracks with TV episodes; (ii) extract human-annotated ADs from AudioVault soundtracks and transcribe them into text forms.

As noted by~\cite{Han24}, the alignments between movies and AudioVault soundtracks are non-trivial owing to (i) non-identical soundtracks; (ii) different recording speeds (e.g.\ NTSC $29.97$ fps \emph{vs.} RAL $25$ fps). Furthermore, in TV series, ADs are typically shorter than those in movies, and there is more frequent mixing between spoken ADs and sound effects (e.g.\ laughing). Both factors pose greater challenges for accurate AD extraction from the soundtracks.

To solve these challenges, we propose an improved pipeline that starts with AudioVault and TV soundtracks and consists of two major steps, namely soundtrack pre-processing and AD filtering, as demonstrated in~\cref{fig:ad_align}.

\vspace{3pt}
\noindent \textbf{Step 1: Soundtrack Preprocessing.} In this step, we convert both AudioVault and TV soundtracks into text-form transcripts and temporally align them. In detail, we first use WhisperX~\cite{Bain23} to transcribe both soundtracks into text forms along with the corresponding timestamps. Additionally, we apply a diarisation module to identify speakers for the AudioVault transcripts. Formally, the resultant transcripts are denoted as $\mathcal{S}_{\text{AV}} = \{(s_1,t_1,c_1), \ldots (s_n,t_n,c_n)\}$ and  $\mathcal{S}_{\text{TV}} = \{(s'_1,t'_1), \ldots (s'_m,t'_m)\}$, where $s_i$ represents transcribed sentences, along with the time stamps $t_i$, and $c_i$ indicates the speaker IDs estimated by the diarisation module. 

These transcribed sentences are then temporally matched in terms of the timestamps by an audio-audio alignment. This process follows the same design as proposed in~\cite{Han24}. (Note, we found audio-audio alignment alone is already precise enough, therefore skipping the first text-text alignment step in their original proposal). Specifically, the soundtracks are transformed into mel-spectrograms ($\mathcal{M}_{\text{AV}}$ and $\mathcal{M}_{\text{TV}}$) that capture the low-level audio representations. Based on the diarisation results, we mask out the regions in the AudioVault mel-spectrogram that overlap with AD timestamps. We then evaluate the correlation map $\mathcal{M}_{\text{AV}}$ and $\mathcal{M}_{\text{TV}}$, and fit it with a linear function $f$ using RANSAC. This function can be used to convert timestamps from AudioVault to TV soundtracks, i.e.\ $f$:$\{t_{\text{AV}} \rightarrow t_{\text{TV}}\}$, therefore aligning two modalities. 

\vspace{3pt}
\noindent \textbf{Step 2: AD Filtering.} As previously noted, simple diarisation on AudioVault soundtracks for TV series can not accurately determine AD speakers with non-AD speakers, and therefore cannot be directly applied to extract ADs. We instead formulate a multi-step filtering mechanism that identifies ADs from ASR transcripts that contain both AD and non-AD speakers. 

First, we compare the pairs of AudioVault sentences $s_t$ (either AD or non-AD speakers) and TV sentences (just non-AD speakers) $s'_{t\pm\Delta t}$, with $\Delta t$ denoting the maximum difference between their timestamp centers. If the word error rate (WER) between a pair of sentences is below a threshold $\epsilon'$, i.e.\ WER$(s_t, s'_{t\pm\Delta t}) < \epsilon'$, we consider the two sentences are matched
(therefore correspond to a non-AD speaker), and remove the sentence $s_t$ from the AudioVault transcripts. As exemplified in~\cref{fig:ad_align}, this filtering by text-text-matching effectively removes a majority of non-AD contents, though still affected by transcription and alignment errors.

Second, we perform diarisation-based filtering that first determines the AD speaker by majority voting, followed by the removal of all sentences from non-AD speakers. After filtering, a few non-AD sentences may remain, albeit marginally, due to potential diarisation errors.

Finally, we conduct rule-based heuristic filtering, which discards sentences containing certain symbols (e.g.\ \textit{``?''} and \textit{``!''}), pronouns (e.g.\ \textit{``I''} and \textit{``We''}), and phrases (e.g.\ \textit{``Let's go.''}), that never appear in ADs. Further details are discussed in~\Cref{supsec:imple}.

%% file: sec/06-Experiments.tex
\section{Experiments}
\label{sec:exp}

\subsection{Datasets}
We evaluate the model performance using several datasets, focusing on character recognition (as an intermediate task), movie AD generation, and TV AD generation.

\vspace{3pt}
\noindent \textbf{Character Recognition Dataset.} For character recognition evaluation, we follow~\cite{Han23a} and adopt a 4-movie subset of \textbf{MovieNet}~\cite{MovieNet}, where ground-truth character names are provided in each keyframe (shot) within the movie.

\vspace{3pt}
\noindent \textbf{Movie AD Datasets.} We report the performance on two movie AD datasets, namely \textbf{CMD-AD}~\cite{Han24} and \textbf{MAD-Eval}~\cite{Han23}. The former is generated by aligning AD data with video clips from CMD (Condensed Movie Dataset)~\cite{Bain20}, resulting in $101$k ground truth ADs within $\sim1.4$k movies. In this work, we adopt its evaluation set (CMD-AD-Eval), which consists of $\sim7$k ADs for $100$ movies. On the other hand, MAD-Eval is a $10$-movie subset from LSMDC~\cite{Rohrbach2017}, featuring $\sim6.5$k ground truth AD annotations. 

\vspace{3pt}
\noindent \textbf{TV AD Dataset.} We adopt the new \textbf{TV-AD} dataset (introduced in~\cref{sec:tvad_dataset}) and measure the performance on its evaluation split, which includes around $3$k ADs in $100$ TV episodes.

\subsection{Evaluation Metrics}
\label{subsec:metric}

To measure the \textbf{character recognition performance} in the intermediate stage, we compare the predicted and ground truth character name lists and compute the IoU, precision, and recall of the prediction.

For AD evaluation, we monitor the performance on three metrics, namely CIDEr, CRITIC, and LLM-AD-Eval.

\vspace{3pt}
\noindent \textbf{CIDEr}~\cite{Vedantam_2015_CVPR} is a common metric that evaluates the quality of text descriptions. It measures the relevance and uniqueness of words in candidate descriptions based on a Term Frequency-Inverse Document Frequency (TF-IDF) weighting scheme. However, this word-matching-based metric has its limitations, as ADs with the same semantics can be presented in various ways. Since many recent works have adopted CIDEr scores, we report the performance for comparison purposes.

\vspace{3pt}
\noindent \textbf{CRITIC}~\cite{Han24} measures the accuracy of character identification in the predicted ADs. Specifically, a co-referencing model is employed to replace ambiguous pronouns such as \textit{``He''} or \textit{``She''} in ADs with official names from the character banks. Then, two name sets from each pair of predicted and GT ADs are compared, and the IoU is calculated to give a CRITIC score.

\vspace{3pt}
\noindent \textbf{LLM-AD-Eval}\cite{Han24} utilises LLMs to judge the quality of generated ADs by scoring them between 1 (lowest) and 5 (highest). The evaluation focuses on the matching between generated and ground truth ADs in terms of human actions, objects, and interactions. We directly adopt the LLM prompt provided in the original paper and use open-source models LLaMA2-7B~\cite{touvron2023llama2openfoundation} and LLaMA3-8B~\cite{llama3modelcard} for the evaluation.

\subsection{Implementation Details}
\label{subsec:imple}
\vspace{3pt}
\noindent \textbf{Character Recognition Details.} To formulate the character bank for each movie or TV episode, we extract top-$10$ characters from the IMDb website\footnote{\url{https://www.imdb.com}} along with their corresponding actor profile images. For face detection and recognition, we adopt the InsightFace package\footnote{\url{https://github.com/deepinsight/insightface}}, which is developed based on ArcFace~\cite{Deng_2019_CVPR} and RetinaFace~\cite{Deng_2020_CVPR}. The threshold for face matching (detailed in~\ref{subsec:face_det}) is set to $\epsilon=0.2$.

\vspace{3pt}
\noindent \textbf{AD Prediction Details.} 
For stage I, we use VideoLLaMA2-7B~\cite{damonlpsg2024videollama2} 
% (with Mistral-7B-Instruct-v0.2~\cite{jiang2023mistral7b} as the language decoder) 
as the Video-LLM model, which takes in $8$ video frames. At stage II, we adopt LLaMA3-8B~\cite{llama3modelcard} for AD summary. The exact text prompts are available in~\Cref{supsec:prompt}.

\begin{table}[t]
\centering
\setlength\tabcolsep{8pt}
\resizebox{0.6\textwidth}{!}{
\begin{tabular}{cccc}  
\toprule
Methods &  IoU & Precision  &  Recall  \\
\midrule
AutoAD-II~\cite{Han23a} & $70.8$ & $75.8$ & $85.6$   \\
AutoAD-Zero \textbf{(Ours)} & $\mathbf{75.8}$ & $\mathbf{79.3}$  & $\mathbf{85.8}$  \\
\bottomrule
\end{tabular}}
\vspace{0.15cm}
\caption{\textbf{Character recognition results} reported on $4$ MovieNet movies. Note, AutoAD-III~\cite{Han24} shares the same character recognition module as AutoAD-II~\cite{Han23a}.
}
\label{tab:char_recog}
\end{table}

\subsection{Character Recognition Results}
In~\cref{tab:char_recog}, we compare the character recognition module with the one adopted in AutoAD-II~\cite{Han23a}, which trains a transformer-based model with CLIP feature inputs to predict the probability of each character. In contrast, our method first employs an off-the-shelf face detection model with a face-matching mechanism to obtain raw character predictions. These predictions are then encoded into coloured circles to visually prompt the Video-LLM, as detailed in~\cref{fig:char_prompt}.

Following~\cite{Han23a}, we report the character recognition performance on $4$ MAD-Eval movies with keyframe character annotations from MovieNet. Note that our reported performance is measured on the final Video-LLM outputs. As demonstrated in~\cref{tab:char_recog}, AutoAD-Zero achieves higher IoU, precision, and recall in character recognition, indicating that our predicted character names are more accurate and reliable compared to those predicted by AutoAD-II.

\subsection{Audio Description Predictions}
\label{subsec:result_ad_pred}
In this section, we compare our training-free method with prior models and provide qualitative examples. We also present a detailed ablation analysis in~\Cref{supsec:abla}, which verifies the key designs in our two-stage framework.

\begin{table}[t!]
\centering
\setlength\tabcolsep{1pt}
\resizebox{\textwidth}{!}{
\begin{tabular}{ccccccccc}  
\toprule
\multirow{2}[2]{*}{Models} & \multirow{2}[2]{*}{\;\shortstack{Training \\ -free}\;} &  \multicolumn{3}{c}{CMD-AD} & MAD-Eval & \multicolumn{3}{c}{TV-AD}  \\
\cmidrule(r){3-5}
\cmidrule(r){6-6}
\cmidrule(r){7-9}
 &  & CIDEr & CRITIC & LLM-AD-eval  & CIDEr &  CIDEr & CRITIC & LLM-AD-eval \\
\midrule
AutoAD-I~\cite{Han23} & \xmark & $-$ & $-$ & $-$ & $14.3$ & $-$ &$-$ & $-$ \\
AutoAD-II~\cite{Han23a} & \xmark & $13.5$ & $8.2$ & $-$ & $19.2$ & $-$ &$-$ & $-$  \\
AutoAD-III~\cite{Han24} & \xmark & $\mathbf{25.0}$ &  $32.7$ & $\mathbf{2.89}$\;|\;$\mathbf{2.01}$  & $24.0$  & $\mathbf{26.1}^*$ & $15.9^*$ & $2.78^*$\;|\;${1.99}^*$ \\
Uni-AD~\cite{wang2024contextual} & \xmark & $-$ & $-$ & $-$ & $\mathbf{28.2}$ & $-$ &$-$ & $-$ \\
\midrule
MM-Narrator (GPT-4)~\cite{Zhang_2024_CVPR} & $\checkmark$  & $-$ & $-$ & $-$  & $13.9$  &$-$ & $-$ & $-$\\
MM-Narrator (GPT-4v)~\cite{Zhang_2024_CVPR} & $\checkmark$  & $-$ & $-$ & $-$  & $9.8$  &$-$ & $-$ & $-$\\
{LLM-AD}~\cite{chu2024llmadlargelanguagemodel} & $\checkmark$  & $-$ & $-$ & $-$  & $20.5$  &$-$ & $-$ & $-$\\
AutoAD-Zero \textbf{(Ours)} & $\checkmark$ & $17.7$ & $\mathbf{43.7}$ & $2.83$\;|\;$1.96$  & $22.4$ & $22.6$ & $\mathbf{27.6}$ & $\mathbf{2.94}$\;|\;$\mathbf{2.00}$ \\
\bottomrule
\end{tabular}}
\vspace{0.15cm}
\caption{\textbf{Comparison with other methods on CMD-AD, MAD-Eval, and TV-AD.} ``$^*$'' denotes the results of the AutoAD-III model that we reproduced by training it on the TV-AD training split. The LLM-AD-eval scores are evaluated using LLaMA2-7B (left) and LLaMA3-8B (right). 
}
\label{tab:quantitative}
\end{table}

% \noindent \textbf{Comparison with Other Methods.} 
\vspace{3pt}
\noindent \textbf{Quantitative Comparison.}
Overall, AD generation methods are majorly divided into two classes: (i) models explicitly trained on ground truth movie or TV AD annotations, including AutoAD-I to III~\cite{Han23,Han23a,Han24} and Uni-AD~\cite{wang2024contextual}; and (ii) \emph{training-free} approaches that perform \emph{zero-shot} AD generation \emph{without} finetuning on AD datasets, including MM-Narrator~\cite{Zhang_2024_CVPR}, {LLM-AD}~\cite{chu2024llmadlargelanguagemodel}, and AutoAD-Zero (our method). The latter methods are more flexible and extendable in terms of performance, as they could be directly integrated with more advanced models. However, it is challenging for training-free models to outperform those fine-tuned on human-annotated ADs that directly benefit from domain-specific knowledge.

We focus on evaluating model performance on CMD-AD and TV-AD. Since several prior works report their CIDEr scores on MAD-Eval, we also include it in our comparison. 
As demonstrated in~\cref{tab:quantitative}, among the training-free methods, AutoAD-Zero surpasses MM-Narrator (both GPT-4 and GPT-4V versions) on the MAD-Eval evaluation. When compared to models fine-tuned on AD datasets, AutoAD-Zero delivers competitive results on CIDEr and LLM-AD-eval metrics and achieves state-of-the-art performance on CRITIC score, indicating our superior character identification accuracy.

\begin{figure}[t!]
    \centering
    \includegraphics[width=0.99\linewidth]{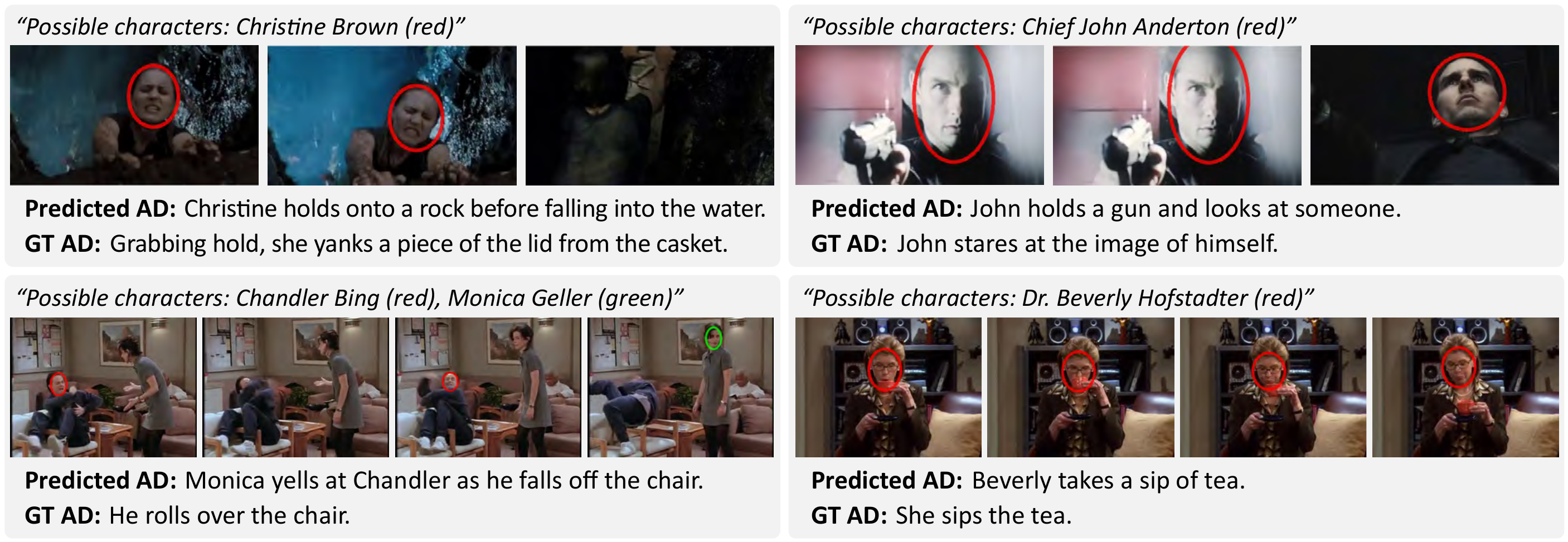}
    \vspace{-0.15cm}
    \caption{\textbf{Qualitative visualisations} for CMD-AD (top) and TV-AD (bottom). The faces are labelled by coloured circles, with corresponding names and colour codes listed above. The examples are from \textit{``Drag Me to Hell''} (top left), \textit{``Minority Report''} (top right), \textit{``Friends''} (bottom left), and \textit{``The Big Bang Theory''} (bottom right).}
    \label{fig:visualisation}
\end{figure}

\vspace{3pt}
\noindent \textbf{Qualitative Visualisation.} \cref{fig:visualisation} provides several examples on CMD-AD and TV-AD. By prompting VLM with character information (i.e.\ coloured circles in images and names in text prompts), AutoAD-Zero is capable of associating characters with their actions. In the bottom left example, even though faces are sparsely labelled (as characters may turn away from the camera), the VLM successfully links the character across the frames (tracking) and identifies individual actions (e.g.\ \textit{Monica - yelling}, \textit{Chandler - falling off the chair}).

%% file: sec/07-Conclusion.tex
\section{Discussion and Extensions}
\label{sec:conclusion}

In this paper, we propose AutoAD-Zero, a zero-shot AD generation framework featuring  visual prompting of character information and a two-stage description-summarisation process. 
% Evaluations on the newly formulated TV-AD dataset and existing movie AD datasets demonstrate that AutoAD-Zero outperforms all training-free methods and effectively competes with models fine-tuned on AD datasets.
The two-stage design offers notable advantages: 
it brings the extensibility to use more advanced VLMs and LLMs in the future, and the flexibility to tailor the second-stage LLM to fit specific styles.

However, the visual and text prompt design is still based on a heuristic trial-and-error approach which could be optimised. Also, errors accumulate between stages -- if the first-stage VLM fails to describe an action, the second-stage LLM cannot summarise it. 
Moreover, recent VLMs and LLMs may have already encountered popular characters and scenes during training, potentially undermining the zero-shot nature of the task. In the future, data leakage could be alleviated by adopting more recent or amateur film data~\cite{ghermi2024shortfilmdatasetsfd} for evaluation.

Potential extensions of this work include: (i) Enhancing the VLM with additional knowledge from ``experts'' or ``specialists'' (e.g.\ location from a scene recognition model); (ii) Expanding the second stage by providing the LLM with more context, including previous ADs, dialogues, and plot summaries.

%% file: sec/Acknowledges.tex
\subsubsection{Acknowledgments.}
This research is supported by the UK EPSRC Programme Grant Visual AI (EP/T028572/1), a Clarendon Scholarship, a Royal Society Research Professorship RP$\backslash$R1$\backslash$191132, and ANR-21-CE23-0003-01 CorVis.

%% file: sec/99-Appendix.tex
\appendix
\begin{center}
    {\large \textbf{Appendix}}
\end{center}
\vspace{0.3cm}
\renewcommand{\thefigure}{\thesection.\arabic{figure}} % \thesection instead of A would make it A.1, B.1...
\setcounter{figure}{0} 
\renewcommand{\thetable}{\thesection.\arabic{table}}
\setcounter{table}{0} 

\noindent This Appendix contains the following parts:
\begin{itemize}
\vspace{-0.1cm}
\setlength\itemsep{0.1em}
    \item \textbf{\cref{supsec:imple}: Additional Implementation Details,} where we provide more information regarding the AD filtering process in TV-AD formulation.
    \item \textbf{\cref{supsec:abla}: Additional Ablation Analysis,} where we investigate the importance of character prompting strategies and factors in AD summary.
    \item \textbf{\cref{supsec:tvad}: Details of TV-AD,} where the detailed TV series with corresponding seasons and episodes are outlined.
    \item \textbf{\cref{supsec:quanti}: More Quantitative Results} are reported on the MAD-Eval dataset.
    \item \textbf{\cref{supsec:quali}: More Qualitative Results} are provided, including the dense video description, predicted ADs, and ground truth ADs.
    \item \textbf{\cref{supsec:prompt}: Detailed Text Prompts} are listed for the VLM and LLM in the two-stage process. The text prompt for LLM-based AD evaluation is also provided.
\end{itemize}
\vspace{-0.1cm}

\section{Additional Implementation Details}
\label{supsec:imple}
\vspace{3pt} \noindent \textbf{Details of Rule-Based AD Filtering.} As mentioned in~\cref{subsec:dataset_formulation}, we adopt a rule-based heuristic approach as the final AD filtering (Step 2). Specifically, a sentence in the transcripts is considered a non-AD sentence if it contains any of the following strings.
\input{fig/rule_based_ad_filtering}

\vspace{3pt} \noindent \textbf{Inference Time Analysis.}
Adopting a two-stage framework in our method allows for more flexible outputs and directly benefits from the improving descriptive power of new VLMs/LLMs. On the other hand, this two-stage design does result in a longer inference time ($0.67$s per AD, with $0.50$s for stage I and $0.16$s for stage II) compared to a single-stage model (e.g.\ $0.20$s for AutoAD-III~\cite{Han24}).

\section{Additional Ablation Studies}
\label{supsec:abla}

\begin{figure}[t!]
    \centering
    \includegraphics[width=0.98\linewidth]{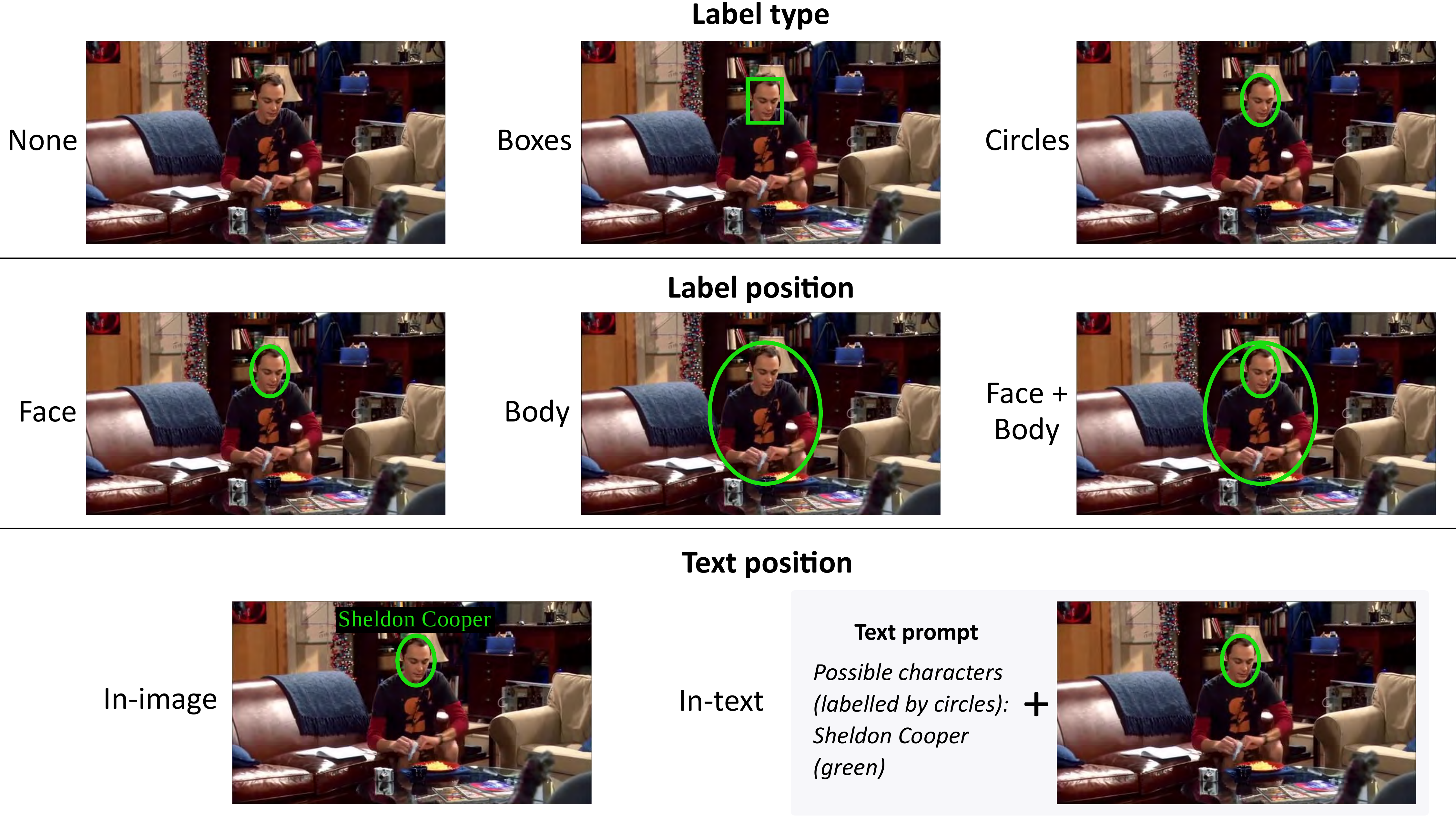}
    \vspace{-0.2cm}
    \caption{\textbf{Schematic visualisations on different character prompting strategies}. ``Label type'' denotes the types of visual indications (e.g.\ boxes or circles). ``Label position'' specifies where the labels are put (e.g.\ around faces or the entire body). ``Text position'' indicates where character names either visually overlaid in images (``In-image'') or provided in text prompts (``In-text'').}
    \label{supfig:char_prompt_demo}
    % \vspace{-0.2cm}
\end{figure}

\begin{table}[hbpt!]
\setlength\tabcolsep{9pt}
\resizebox{\textwidth}{!}{
\begin{tabular}{cccccccc}  
\toprule
\multirow{2}[2]{*}{Exp.} & \multirow{2}[2]{*}{\shortstack{Label \\ type}} & \multirow{2}[2]{*}{\shortstack{Label \\ position}}  & \multirow{2}[2]{*}{\shortstack{Text \\ position}} & \multicolumn{2}{c}{CMD-AD} & \multicolumn{2}{c}{TV-AD} \\
\cmidrule(r){5-6}
\cmidrule(r){7-8}
 & & & & CIDEr & CRITIC & CIDEr & CRITIC \\
\midrule
A & \cellcolor{lightgray}None  & \cellcolor{lightgray}$-$  & \cellcolor{lightgray}$-$ & $9.2$ & $0.8$ & $13.5$ &  $0.1$  \\
B &  \cellcolor{lightgray}None  & \cellcolor{lightgray}$-$ &  In-text & $15.8$ & $34.7$ & $21.0$ &  $27.4$ \\
C & \cellcolor{lightgray}None & \cellcolor{lightgray} $-$ & \cellcolor{lightgray}In-image  & $15.1$  & $35.0$ & $20.9$ & $24.2$  \\
\midrule
D & \cellcolor{lightgray}Boxes & Face & \cellcolor{lightgray}In-image & $15.2$ & $32.5$ &  $19.5$ & $21.1$   \\
E & Circles & Face & \cellcolor{lightgray}In-image  & $15.2$ & $36.6$ & $19.7$ & ${21.1}$  \\
\midrule
F & \cellcolor{lightgray}Boxes & Face & In-text  &  $17.3$ & $42.6$ & ${22.3}$ &  $24.9$ \\
G (default) & Circles & Face & In-text & $\mathbf{17.7}$ & $\mathbf{43.7}$ & $\mathbf{22.6}$ & ${27.6}$ \\
\midrule
H & \cellcolor{lightgray}Boxes & \cellcolor{lightgray}Body & In-text  &  $17.3$ & $39.3$ & ${21.6}$ &  $26.3$ \\
I & Circles & \cellcolor{lightgray}Body & In-text & ${17.2}$ & ${40.7}$ & ${22.3}$ & $\mathbf{28.3}$ \\
\midrule
J & \cellcolor{lightgray}Boxes & \cellcolor{lightgray}Face+Body & In-text  &  $17.2$ & $41.1$ & ${21.0}$ &  $24.8$ \\
K & Circles & \cellcolor{lightgray}Face+Body & In-text & $16.6$ & $42.2$ & $20.2$ & ${26.8}$ \\
\bottomrule
\end{tabular}}
\vspace{0.15cm}
\caption{\textbf{Comparison across different character prompting strategies.} We highlight the changes from the default setting (Exp. G) with gray. Visualisations on different label types, label positions, and text positions are available in~\cref{supfig:char_prompt_demo}.}
\label{suptab:char_prompt_ablation}
\vspace{-0.5cm}
\end{table}

\vspace{3pt} \noindent \textbf{Character Prompting Strategies.} 
\cref{supfig:char_prompt_demo} illustrates several strategies for visually prompting the VLM with character information, divided into several aspects:  (i) Label types, which can be in the form of boxes or circles (or no labels at all); (ii) Label position, which could be around the person's face and/or around the whole body; (iii) Text position indicating where the character names are provided, either "in-image" or "in-text". The former relies on the OCR capability of the VLM and asks it to extract the names from the image, while the latter simply provides the names in the text 
prompt, associated with corresponding circles or boxes in the image by colour codes. 

Several combinations of these strategies are investigated, with results reported in~\cref{suptab:char_prompt_ablation}. The following observations can be made: 
\begin{itemize}
    \setlength\itemsep{0.3em}
    \item \textit{Exp.\,A vs.\ Exp.\,G:\;} If no character information is offered, the resultant scores are significantly lower.
    \item \textit{Exp.\,B vs.\ Exp.\,G:\;} Exp.\,B corresponds to the case where character names are only provided in text prompts without any visual indications. This approach works in scenarios with only one character but can cause mis-associations when multiple people are present, leading to degraded results.
    \item \textit{Exp.\,C vs.\ Exp.\,G:\;} {Exp.\,C indicates that character names are directly overlaid in the image (above each character's head), without any additional labelling (e.g.\ circles or boxes). This setting is similar to the visual prompting used in the concurrent LLM-AD~\cite{chu2024llmadlargelanguagemodel}. However, it does not lead to better performance than our default setting, possibly due to errors introduced in the additional OCR step for the VLM.}
    \item \textit{Exp.\,E vs.\ Exp.\,G:\;} Directly providing the character names in text prompts (``In-text'') is preferable to overlaying names in images (``In-image'')
    \item \textit{Exp.\,F vs.\ Exp.\,G:\;} Adopting circles as visual character indicators slightly outperforms the box counterparts. 
    \item \textit{Exp.\,I\&K vs.\ Exp.\,G:\;} Applying labels around the human body achieves similar performance to using labels around faces. However, when both face and body labels are used simultaneously, performance degrades, possibly owing to the confusion caused by multiple labels indicating the same identity.    
\end{itemize}
% \vspace{-0.2cm}
As a result, we apply ``Circle\,+\,Face\,+\,In-text'' (Exp.\,F) as our default setting.

\vspace{10pt}
\noindent \textbf{Factors for Video Description.} 
\cref{tab:factor_ablation} outlines several factors considered for VLM-based video description (Stage I) and demonstrates their influences on the final AD results. Specifically, if the characters' actions and interactions are removed from the description (rows 1 and 2), both CIDEr and CRITIC scores witness significant drops, suggesting the crucial roles of these two factors in capturing the dynamics in the video. If facial expressions are not taken into account (row 3), the character identification would be affected, leading to a decrease in CRITIC score.

In addition to actions, interactions, and facial expressions, three extra factors are investigated, namely environments (row 4), character appearances (row 5), and interactions with objects (row 6). As can be observed, adding these factors adversely affects the performance. We suspect this could be because such information is not necessary for \emph{every} AD and may be partially covered by factors such as actions and interactions. As a result, providing excessive information to the second stage could make it more difficult for the LLM to summarise the most important details for AD generation, resulting in degraded performance.

\vspace{10pt} \noindent \textbf{Factors for AD summary.} 
During the second stage in our framework, we use an LLM to summarise the dense video description into a one-sentence AD output. As detailed in Sec. 4.2 of the main text, multiple factors are considered when formulating the prompt instructions, including rules for character and action descriptions, adjustment of AD lengths, and providing AD examples. As demonstrated in~\cref{suptab:factor_ablation}, performance degrades when any of these factors are removed. In particular, when no ground truth AD examples are provided (row 4 in~\cref{suptab:factor_ablation}), both CIDEr and CRITIC scores drop significantly.

\begin{table}[hbpt!]
\centering
\vspace{-0.2cm}
\setlength\tabcolsep{4pt}
\resizebox{\textwidth}{!}
{
\begin{tabular}{cccccccccc}   
\toprule
\multicolumn{6}{c}{Factors for video descriptions} &  \multicolumn{2}{c}{CMD-AD} & \multicolumn{2}{c}{TV-AD}  \\
\cmidrule(r){1-6}
\cmidrule(r){7-8}
\cmidrule(r){9-10}
Action & Interaction & Facial exp.  & Environ. & Appear. & Object & CIDEr & CRITIC & CIDEr & CRITIC \\
\midrule
\xmark & \cellcolor{lightgreen}$\checkmark$ & \cellcolor{lightgreen}$\checkmark$ & \xmark  & \xmark & \xmark & $15.7$ & $42.0$ & $18.9$ & $26.4$\\
\cellcolor{lightgreen}$\checkmark$ & \xmark & \cellcolor{lightgreen}$\checkmark$ &  \xmark  & \xmark & \xmark & $16.4$ & $43.1$ & $19.9$ & $27.0$\\
\cellcolor{lightgreen}$\checkmark$ & \cellcolor{lightgreen}$\checkmark$ & \xmark &  \xmark  & \xmark & \xmark & $16.7$ & $38.6$ & $20.6$ & $26.1$\\
\midrule
\cellcolor{lightgreen}$\checkmark$ & \cellcolor{lightgreen}$\checkmark$ & \cellcolor{lightgreen}$\checkmark$ & \cellcolor{lightgreen}$\checkmark$  & \xmark  & \xmark & $16.5$ & $38.6$ & $22.1$ & $25.8$\\
\cellcolor{lightgreen}$\checkmark$  & \cellcolor{lightgreen}$\checkmark$ & \cellcolor{lightgreen}$\checkmark$ & \xmark & \cellcolor{lightgreen}$\checkmark$  & \xmark  & $15.5$ & $37.0$ & $21.3$ & $27.3$\\
\cellcolor{lightgreen}$\checkmark$  & \cellcolor{lightgreen}$\checkmark$ & \cellcolor{lightgreen}$\checkmark$ & \xmark   & \xmark & \cellcolor{lightgreen}$\checkmark$  & $17.5$ & $42.5$ & $21.3$ & $26.7$\\
\midrule
\cellcolor{lightgreen}$\checkmark$  & \cellcolor{lightgreen}$\checkmark$ & \cellcolor{lightgreen}$\checkmark$ & \xmark & \xmark  & \xmark  & $\mathbf{17.7}$ & $\mathbf{43.7}$ & $\mathbf{22.6}$ & $\mathbf{27.6}$\\
\bottomrule
\end{tabular}
}
\vspace{0.15cm}
\caption{\textbf{Ablations on the factors for describing videos.} Our default configuration considers actions, interactions, and facial expressions as the main factors for video description.}
\label{tab:factor_ablation}
\vspace{-0.9cm}
\end{table}

\begin{table}[hbpt!]
\centering
\vspace{-0.4cm}
\setlength\tabcolsep{8pt}
\resizebox{\textwidth}{!}
{
\begin{tabular}{cccccccc}   
\toprule
\multicolumn{4}{c}{Factors for AD summary} &  \multicolumn{2}{c}{CMD-AD} & \multicolumn{2}{c}{TV-AD}  \\
\cmidrule(r){1-4}
\cmidrule(r){5-6}
\cmidrule(r){7-8}
\multirow{2}{*}{\shortstack{Character \vspace{0.02cm} \\ description}} & \multirow{2}{*}{\shortstack{Action  \vspace{0.1cm}\\  summary}}  &  \multirow{2}{*}{\shortstack{Length \\ adjustment}}  &  \multirow{2}{*}{Examples} & \multirow{2}{*}{CIDEr} & \multirow{2}{*}{CRITIC} & \multirow{2}{*}{CIDEr} & \multirow{2}{*}{CRITIC} \\
 &  &  &  &  &  &  &  \\
\midrule
\xmark & \cellcolor{lightgreen}$\checkmark$ & \cellcolor{lightgreen}$\checkmark$ & \cellcolor{lightgreen}$\checkmark$   & $15.5$ & $29.9$ & $20.4$ & $25.9$\\
\cellcolor{lightgreen}$\checkmark$ & \xmark & \cellcolor{lightgreen}$\checkmark$ &  \cellcolor{lightgreen}$\checkmark$  & $16.8$ & $41.2$ & $20.4$ & $27.1$\\
\cellcolor{lightgreen}$\checkmark$ & \cellcolor{lightgreen}$\checkmark$ & \xmark &  \cellcolor{lightgreen}$\checkmark$  & $17.4$ & $41.8$ & $22.0$ & $26.9$\\
\cellcolor{lightgreen}$\checkmark$ & \cellcolor{lightgreen}$\checkmark$ &  \cellcolor{lightgreen}$\checkmark$  & \xmark & $13.7$ & $36.9$ & $15.5$ & $26.5$\\
\midrule
\cellcolor{lightgreen}$\checkmark$ & \cellcolor{lightgreen}$\checkmark$ & \cellcolor{lightgreen}$\checkmark$ & \cellcolor{lightgreen}$\checkmark$   & $\mathbf{17.7}$ & $\mathbf{43.7}$ & $\mathbf{22.6}$ & $\mathbf{27.6}$ \\
\bottomrule
\end{tabular}
}
\vspace{0.1cm}
\caption{\textbf{Ablations on the factors for AD summary.} Detailed explanations regarding these factors can be found in Sec. 4.2 in the main text. The exact text prompts can be found in Sec. E.}
\label{suptab:factor_ablation}
\vspace{-0.55cm}
\end{table}

\vspace{3pt} \noindent \textbf{Change of Question Order in Video Description.}
We permute the question orders in the video description process (stage I). The results are reported in~\cref{suptab:question_order}, which demonstrates the importance of asking the model to identify the characters as the first step for the chain-of-thought process.
\begin{table}[h!]
\vspace{-0.4cm}
\centering
\setlength\tabcolsep{12pt}
\resizebox{0.84\textwidth}{!}{
\begin{tabular}{ccccc}  
\toprule
\multirow{2}[2]{*}{\shortstack{Question \\ order}}  &  \multicolumn{2}{c}{CMD-AD} & \multicolumn{2}{c}{TV-AD}  \\
\cmidrule(r){2-3}
\cmidrule(r){4-5}
 &  CIDEr & CRITIC &  CIDEr & CRITIC \\
\midrule
\textcolor{blue}{\textbf{C}}-I-A-F & $18.1$  & $41.2$ & $22.2$  &  $26.5$  \\
\textcolor{blue}{\textbf{C}}-F-A-I & $17.1$ & $42.0$ & $20.1$  & $27.9$  \\
I-\textcolor{blue}{\textbf{C}}-A-F & $15.2$ & $28.9$ & $19.1$  & $20.0$  \\
\midrule
\textcolor{blue}{\textbf{C}}-A-I-F (default) & $17.7$ & $43.7$ & $22.6$ & $27.6$ \\
\bottomrule
\end{tabular}}
\vspace{0.15cm}
\caption{\textbf{Change of question ordering in stage I.} C: characters; A: actions; I: interactions; F: facial expressions.}
\label{suptab:question_order}
\vspace{-0.5cm}
\end{table}

\noindent \textbf{Different VLM and LLM Models.} 
We also investigate how the proposed framework generalises across different VLMs and LLMs, with the results provided in~\cref{tab:vlm&llm_ablation}. The first row corresponds to an experiment without the second stage, where a one-sentence AD is directly summarised in the first stage. This leads to a noticeable degradation in performance, verifying the necessity of the LLM-based second stage for AD summary.

In terms of VLMs, the more recent VideoLLaMA2 outperforms the earlier Video-LLaVA model. For the LLM adopted for AD summary, LLaMA3 and Gemma2 demonstrate similar performance. The training-free nature of AutoAD-Zero allows it to be easily applied to more advanced VLM and LLM models in the future, which could lead to even better performance.

\begin{table}[hbpt!]
\centering
\vspace{-0.4cm}
\setlength\tabcolsep{8pt}
\resizebox{0.9\textwidth}{!}
{
\begin{tabular}{cccccc}   
\toprule
\multicolumn{2}{c}{Model setting} &  \multicolumn{2}{c}{CMD-AD} & \multicolumn{2}{c}{TV-AD}  \\
\cmidrule(r){1-2}
\cmidrule(r){3-4}
\cmidrule(r){5-6}
Stage I & Stage II & CIDEr & CRITIC & CIDEr & CRITIC \\
\midrule
VideoLLaMA2-7B~\cite{damonlpsg2024videollama2} & \cellcolor{lightgray}$-$ & $12.9$ & $33.2$ & $12.1$ & $26.6$\\
\cellcolor{lightgray}Video-LLaVA-7B~\cite{lin2023video} & LLaMA3-8B~\cite{llama3modelcard} & $15.1$ & $36.4$ & $19.4$ & $\mathbf{28.8}$\\
VideoLLaMA2-7B~\cite{damonlpsg2024videollama2} & \cellcolor{lightgray}Gemma2-9B~\cite{gemma_2024} & $17.3$ & $42.1$ & $21.9$ & $26.6$\\
\midrule
VideoLLaMA2-7B~\cite{damonlpsg2024videollama2} & LLaMA3-8B~\cite{llama3modelcard} & $\mathbf{17.7}$ & $\mathbf{43.7}$ & $\mathbf{22.6}$ & $27.6$\\
\bottomrule
\end{tabular}
}
\vspace{0.15cm}
\caption{\textbf{Different VLM and LLM models.} We highlight the changes from the last row with gray. The first row corresponds to the experiment without the second stage (no AD summary).}
\label{tab:vlm&llm_ablation}
\end{table}

\noindent \textbf{Repeated Experiments.} 
We repeated experiments to account for randomness introduced in the VLM and LLM sampling process. The results shown in \cref{tab:rep_exp} are consistent with relatively low variance, validating the robustness of our experimental findings across ablation studies and the final result.

\begin{table}[tbph!]
\vspace{-0.35cm}
\centering
\setlength\tabcolsep{6pt}
\resizebox{0.9\textwidth}{!}{
\begin{tabular}{cccccccc}  
\toprule
\multirow{2}[2]{*}{Exp.}  &  \multicolumn{3}{c}{CMD-AD} & MAD-Eval & \multicolumn{3}{c}{TV-AD}  \\
\cmidrule(r){2-4}
\cmidrule(r){5-5}
\cmidrule(r){6-8}
 &  CIDEr & CRITIC & LLM-Eval  & CIDEr &  CIDEr & CRITIC & LLM-Eval \\
\midrule
1 & $17.7$ & $43.7$ & $2.83$\;$\mid$\;$1.96$  & $22.4$ & $22.6$ & $27.6$ & $2.94$\;$\mid$\;$2.00$ \\
2 & $17.4$ & $44.5$ & $2.82$\;$\mid$\;$1.98$  & $22.0$ & $22.0$ & $29.5$ & $2.95$\;$\mid$\;$1.95$ \\
3 & $17.6$ & $42.4$ & $2.82$\;$\mid$\;$1.97$  & $22.5$ & $22.3$ & $27.3$ & $2.96$\;$\mid$\;$1.95$ \\
\midrule
mean & $17.6$ & $43.5$ & $2.82$\;$\mid$\;$1.97$  & $22.3$ & $22.3$ & $28.1$ & $2.95$\;$\mid$\;$1.97$ \\
std. & $0.2$ & $1.1$ & $0.01$\;$\mid$\;$0.01$  & $0.2$ & $0.3$ & $1.2$ & $0.01$\;$\mid$\;$0.03$ \\
\bottomrule
\end{tabular}}
\vspace{0.15cm}
\caption{\textbf{Repeated experiments.}}
\label{tab:rep_exp}
\vspace{-0.3cm}
\end{table}

\section{Details of TV-AD}
\label{supsec:tvad}
The TV series in TV-AD training and evaluation splits are listed in~\cref{suptab:tv-ad}, with detailed seasons and episodes provided. The evaluation split will be publicly available, with video sources sampled from TVQA~\cite{lei2018tvqa}. The ground truth AD annotations for both training and evaluation splits will also be released.

\begin{table}[th!]
\centering
\setlength\tabcolsep{10pt}
\resizebox{0.95\textwidth}{!}{
\begin{tabular}{cl}  
\toprule
Split &  TV series in TV-AD (with seasons and episodes) \\
\midrule
Train &  \multirow{10}{*}{\shortstack[l]{\textit{Star Trek: The Next Generation}: S1(E1-E25), S2(E1-E22); \\ \textit{Seinfeld}:  S3(E1-E2, E6-E9, E13-E14), S4(E1-E3, E5-E13, E18-E19, E23); \\ \textit{Frasier}: S1(E1-E24), S2(E1-E2, E5-E24); \\ \textit{Millennium}: S1(E1-E5, E7-E12, E15-E22), S2(E1-E16, E20-E23); \\ \textit{Buffy the Vampire Slayer}: S1(E1-E12), S2(E1-E23), S3(E1-E22); \\ \textit{Battlestar Galactica}: S1(E1-E13); \\
\textit{Desperate Housewives}: S2(E1-E4, E6-E7, E9-E23), S3(E1-E23); \\ \textit{Lost}: S1(E1-E24), S2(E1-E23); \\ \textit{30 Rock}: S5(E1-3); \\ \textit{Breaking Bad}: S1(E1-E7)}}\\
& \\
& \\
& \\
& \\
& \\
& \\
& \\
& \\
& \\
& \\
% \textcolor{gray}{Our face-detection-based model} & \textcolor{gray}{76.9} & \textcolor{gray}{80.9}  & \textcolor{gray}{86.4}  \\
\midrule
Eval &  \multirow{2}{*}{\shortstack[l]{\textit{Friends}: S1(E1-E16, E18-E24), S2(E1-12), S3(E1-E25);\\ \textit{The Big Bang Theory}: S1(E1-E17), S2(E1-E23)}}  \\
& \\
\bottomrule
\end{tabular}}
\vspace{0.15cm}
\caption{\textbf{TV-AD details}, where TV series names are shown along with the adopted seasons and episodes. Note that there are some missing episodes, which are filtered out owing to the misalignment between AudioVault soundtracks and video sources.}
\vspace{-0.6cm}
\label{suptab:tv-ad}
\end{table}

\section{More Quantitative Results}
\label{supsec:quanti}
In \cref{suptab:metric}, we have additionally reported the performance on Rouge-L, METEOR, and BLEU-1, demonstrating our superior performance among training-free methods.
\begin{table}[h!]
\vspace{-0.25cm}
\centering
\setlength\tabcolsep{7pt}
\resizebox{0.9\textwidth}{!}{
\begin{tabular}{cccccc}  
\toprule
\multirow{2}[2]{*}{Models} & \multirow{2}[2]{*}{\shortstack{Training \\ -free}\;} &  \multicolumn{4}{c}{MAD-Eval}  \\
\cmidrule(r){3-6}
 &  & CIDEr &  Rouge-L & METEOR & BLEU-1 \\
\midrule
AutoAD-I & \xmark & $14.3$ & $10.3$ &$-$ & $-$ \\
AutoAD-II & \xmark & $19.2$ & $13.4$ &$-$ & $-$  \\
AutoAD-III & \xmark & $24.0$  & $13.9$ & $\mathbf{5.5}$ & $\mathbf{13.1}$ \\
Uni-AD & \xmark &$\mathbf{28.2}$ & $\mathbf{17.2}$ &$-$ & $-$ \\
\midrule
MM-Narrator (GPT-4) & $\checkmark$ & $13.9$  &$13.4$ & $6.7$ & $12.8$\\
MM-Narrator (GPT-4v) & $\checkmark$  & $9.8$  &$12.8$ & $\mathbf{7.1}$ & $10.9$\\
% \revised{LLM-AD} & $\checkmark$   & $20.5$  &$-$ & $-$ & $-$\\
AutoAD-Zero \textbf{(Ours)} & $\checkmark$  & $\mathbf{22.4}$ & $\mathbf{14.4}$ & $6.6$ & $\mathbf{15.1}$ \\
\bottomrule
\end{tabular}}
\vspace{0.15cm}
\caption{\textbf{Comparison with other methods on MAD-Eval.} Additional metrics such as Rouge-L, METEOR, and BLEU-1 are reported.
}
\label{suptab:metric}
\vspace{-0.8cm}
\end{table}

\section{More Qualitative Results}
\label{supsec:quali}
\begin{figure}[thbp!]
    \centering
    \includegraphics[width=0.99\linewidth]{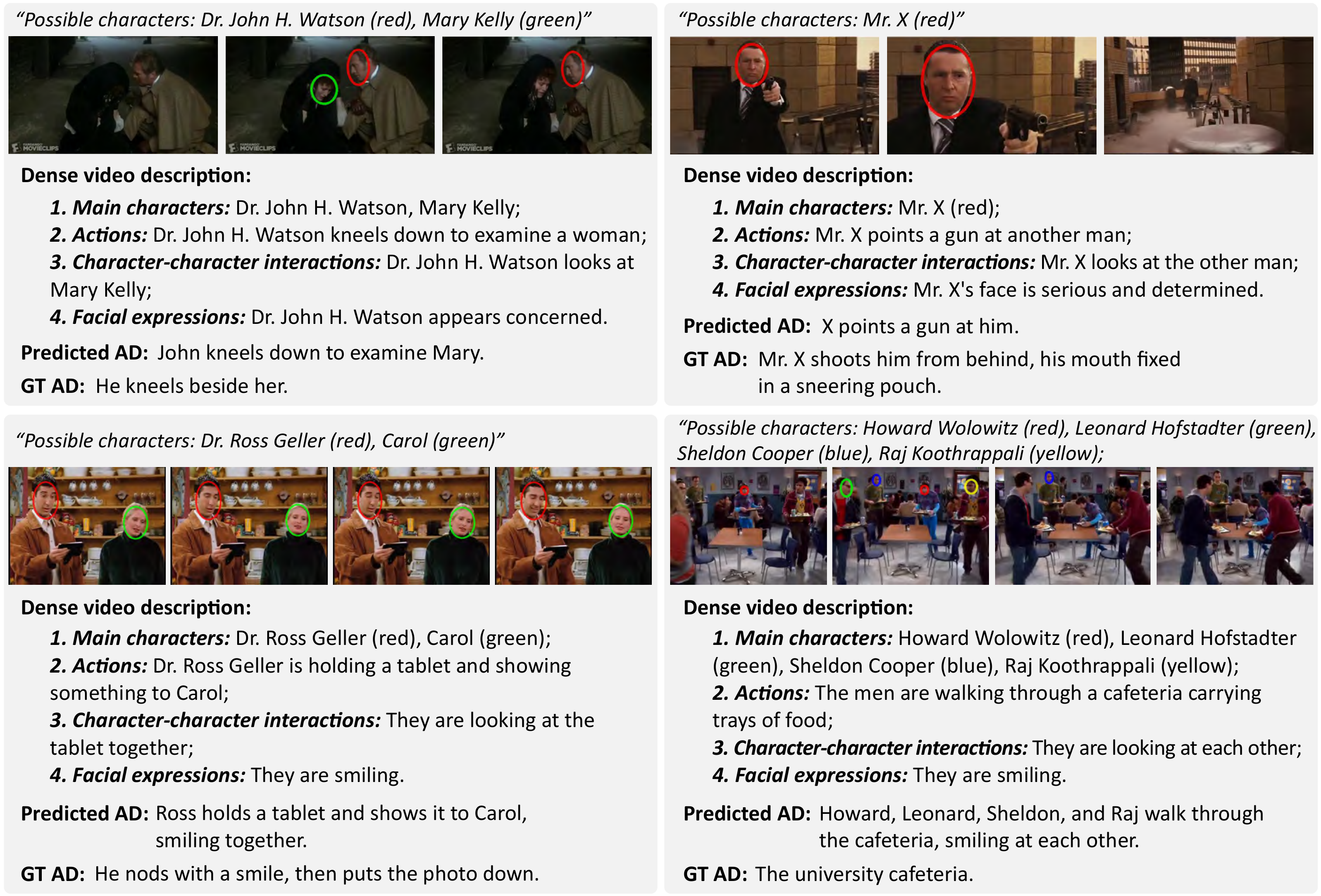}
    \vspace{-0.25cm}
    \caption{\textbf{Additional visualisations} for CMD-AD (top) and TV-AD (bottom). The faces are labelled by coloured circles, with corresponding names and colour codes in text prompts above each video clip. The intermediate outputs from the VLM in stage I (dense video description) are also provided, which are taken as the input for the LLM-based AD summary in stage II. The examples shown are from \textit{``Murder by Decree''} (top left), \textit{``Wanted''} (top right), \textit{``Friends''} (bottom left), and \textit{``The Big Bang Theory''} (bottom right).}
    \label{supfig:visualisation}
    \vspace{-0.3cm}
\end{figure}

Additional qualitative visualisations are provided in~\cref{supfig:visualisation}. Besides the predicted and ground truth ADs, dense video descriptions output from the VLM (in the intermediate stage I) are also included to illustrate how the comprehensive information is summarised into a single AD sentence. 

\begin{figure}[h!tpb]
    \vspace{-0.4cm}
    \centering
    \includegraphics[width=\linewidth]{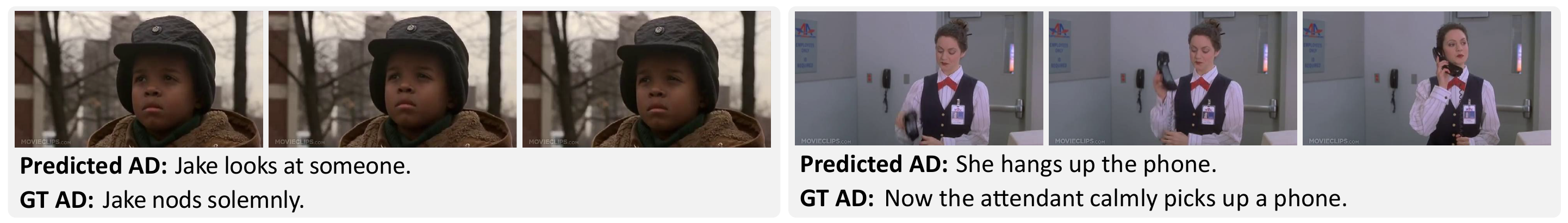}
    \vspace{-0.6cm}
    \caption{\textbf{Failure case visualisation.} The examples shown are from \textit{``Candyman''} (left) and \textit{``Meet the Parents''} (right).}
    % \vspace{-0.2cm}
    \label{supfig:rebuttal_figure_vis}
\end{figure}

\cref{supfig:rebuttal_figure_vis} illustrates two exemplary failure cases, both closely related to limitations in VLM prediction. The first example involves minimal motion (nodding), which the VLM fails to capture. The second example highlights the ``arrow of time'' issue, where the VLM occasionally misinterprets the direction of an action.

Please refer to more visualisations in the project page video, where both predicted and ground truth ADs are provided in text form (i.e.\ ``predicted AD || ground truth AD''). The voice of predicted AD is generated by the OpenAI text-to-speech API\footnote{\url{https://platform.openai.com/docs/guides/text-to-speech}} and added to the original video soundtrack.

\section{Detailed Text Prompts}
\label{supsec:prompt}
The exact text prompts for the two-stage process (including the alternatives used in ablation analysis) are provided in~\cref{supalg:stage1} (Stage I: VLM-based video description) and in~\cref{supalg:stage2} (Stage II: LLM-based AD summary). The detailed prompt for LLM-AD-eval is also available in~\cref{supalg:llmeval}.
\input{fig/exact_prompt_stage1}

\input{fig/exact_prompt_stage2}
\input{fig/exact_prompt_llm_eval}

%% file: fig/rule_based_ad_filtering.tex
% Configuration for the listings package
\definecolor{DeepPurple}{rgb}{0.41, 0.11, 0.69}
\definecolor{DeepGreen}{rgb}{0.0, 0.5, 0.0}
\lstset{ 
  language=python, 
  basicstyle=\ttfamily\scriptsize, 
  keywordstyle=\color{blue}\bfseries,
  commentstyle=\color{DeepGreen},
  stringstyle=\color{DeepPurple},
  showstringspaces=false,
  numberstyle=\tiny\color{gray},
  numbersep=5pt,
  breaklines=true,
  captionpos=b,
  escapeinside={(*@}{@*)}
}
% \vspace{-0.1cm}
\begin{lstlisting}
new_info = info[~info['gt_ad'].str.contains(r'\?|I\'|I |You|you\'|you |We|we\'|we |\!|Oh,|Hey,|Okay,|Yeah,|Um.|me.|Let\'s|you.|Hi,|Yes,|No,|Hello,|Aw,|created by|Created by|Uh,|Directed by|directed by|Written by|written by|Edited by|edited by|Music by|music by|Audio description|A caption|Another caption.|another caption.|Credits appear.|credits appear.|This\'s|That\'s|Alright|A title|Title|Copyright|Good morning|See ya.|See,|All right|Got it.|Mm,|One, two,|Oops,|No problem.|Good night|Ah,|Ooh,|Wow,|Very well.|Good luck.|Good,|Bye-bye,|OK,', regex=True)]
\end{lstlisting}
% \vspace{-0.1cm}

%% file: fig/exact_prompt_stage1.tex
% Configuration for the listings package
\definecolor{DeepPurple}{rgb}{0.41, 0.11, 0.69}
\definecolor{DeepGreen}{rgb}{0.0, 0.5, 0.0}
\lstset{ 
  language=python, 
  basicstyle=\ttfamily\scriptsize, 
  keywordstyle=\color{blue}\bfseries,
  commentstyle=\color{DeepGreen},
  stringstyle=\color{DeepPurple},
  showstringspaces=false,
  numberstyle=\tiny\color{gray},
  numbersep=5pt,
  breaklines=true,
  captionpos=b,
  escapeinside={(*@}{@*)}
}
\begin{algorithm}[htb!]
% \vspace{0.2cm}
\caption{Stage I text prompt for VideoLLaMA2-7B}\label{supalg:stage1}
% \vspace{-0.2cm}
\begin{lstlisting}
prompt = (
    "Please describe the TV series clip in the following four steps: "
    "1. Identify main characters (if {label_type} are available){char_text}; "
    "2. Describe the actions of characters in one sentence, i.e., who is doing what, focusing on the movements; " 
    "3. Describe the interactions between characters in one sentence, such as looking; "
    "4. Describe the facial expressions of characters in one sentence. "
    "Note, colored {label_type} are provided for character indications only, DO NOT mention them in the description. "   
    "Make sure you do not hallucinate information. "
    "###ANSWER TEMPLATE###: 1. Main characters: ''; 2. Actions: ''; 3. Character-character interactions: ''; 4. Facial expressions: ''."
) 

# Optional prompts for environments, appearances and objects, can be insert above if needed.
# "Describe the environments in one sentence, focusing on the location, furniture, entrances and exits, etc.; "
# "Describe the appearances and costumes of characters in one sentence; "
# "Describe the interactions between objects and characters in one sentence; "
\end{lstlisting}
% \vspace{-0.4cm}
\end{algorithm}

%% file: fig/exact_prompt_stage2.tex
% Configuration for the listings package
\definecolor{DeepPurple}{rgb}{0.41, 0.11, 0.69}
\definecolor{DeepGreen}{rgb}{0.0, 0.5, 0.0}
\lstset{ 
  language=python, 
  basicstyle=\ttfamily\scriptsize, 
  keywordstyle=\color{blue}\bfseries,
  commentstyle=\color{DeepGreen},
  stringstyle=\color{DeepPurple},
  showstringspaces=false,
  numberstyle=\tiny\color{gray},
  numbersep=5pt,
  breaklines=true,
  captionpos=b,
  escapeinside={(*@}{@*)}
}
\begin{algorithm}[htb!]
% \vspace{0.2cm}
\caption{Stage II text prompt for LLaMA3-8B}\label{supalg:stage2}
\vspace{-0.2cm}
\begin{lstlisting}
sys_prompt = (
    "[INST] <<SYS>> \n You are an intelligent chatbot designed for summarizing TV series audio descriptions. Here's how you can accomplish the task:------##INSTRUCTIONS: you should convert the predicted descriptions into one sentence. You should directly start the answer with the converted results WITHOUT providing ANY more sentences at the beginning or at the end.\n<</SYS>>\n\n{} [/INST] "
)

# For TV-AD
verb_list = ['look', 'walk', 'turn', 'stare', 'take', 'hold', 'smile', 'leave', 'pull', 'watch', 'open', 'go', 'step', 'get', 'enter']

# For CMD-AD and MAD-Eval
verb_list = ['look', 'turn', 'take', 'hold', 'pull', 'walk', 'run', 'watch', 'stare', 'grab', 'fall', 'get', 'go', 'open', 'smile']

# Insert stage I output "{stage_1_output}" and AD duration "{duration}"
user_prompt = (
    "Please summarise the following description for one TV series clip into ONE succinct audio description (AD) sentence.\n"
    f"Description: {stage_1_output}\n\n"
    "Focus on the most attractive characters and their actions.\n"
    "For characters, use their first names, remove titles such as 'Mr.' and 'Dr.'. If names are not available, use pronouns such as 'He' and 'her', do not use expression such as 'a man'.\n"
    "For actions, avoid mentioning the camera, and do not focus on 'talking' or position-related ones such as 'sitting' and 'standing'.\n"
    "Do not mention characters' mood.\n"
    "Do not hallucinate information that is not mentioned in the input.\n"
    f"Try to identify the following motions (with decreasing priorities): {verb_list}, and use them in the description.\n"
    "Provide the AD from a narrator perspective and adjust the length of the output according to the duration.\n"
    f"Duration of the video clip: {duration}s\n\n"
    "For example, a response of duration 0.8s could be: {'summarised_AD': 'She looks at Riker.'}.\n"
    "Another example response of duration 1.4s is: {'summarised_AD': 'Paul looks at his wife lovingly.'}.\n"
    "An example response of duration 2.6s is: {'summarised_AD': 'He watches Tasha calmly battle with the figure.'}.\n"
)

messages = [
    {"role": "system", "content": sys_prompt},
    {"role": "user", "content": user_prompt},
]
\end{lstlisting}
% \vspace{-0.2cm}
\end{algorithm}

%% file: fig/exact_prompt_llm_eval.tex
% Configuration for the listings package
\definecolor{DeepPurple}{rgb}{0.41, 0.11, 0.69}
\definecolor{DeepGreen}{rgb}{0.0, 0.5, 0.0}
\lstset{ 
  language=python, 
  basicstyle=\ttfamily\scriptsize, 
  keywordstyle=\color{blue}\bfseries,
  commentstyle=\color{DeepGreen},
  stringstyle=\color{DeepPurple},
  showstringspaces=false,
  numberstyle=\tiny\color{gray},
  numbersep=5pt,
  breaklines=true,
  captionpos=b,
  escapeinside={(*@}{@*)}
}
\begin{algorithm}[htb!]
% \vspace{0.2cm}
\caption{Text prompts for LLM-AD-eval on LLaMA2-7B and LLaMA3-8B}\label{supalg:llmeval}
\vspace{-0.2cm}
\begin{lstlisting}
sys_prompt = "[INST] <<SYS>>\nYou are an intelligent chatbot designed for evaluating the quality of generative outputs for movie audio descriptions. Your task is to compare the predicted audio descriptions with the correct audio descriptions and determine its level of match, considering mainly the visual elements like actions, objects and interactions. Here's how you can accomplish the task:------##INSTRUCTIONS: - Check if the predicted audio description covers the main visual events from the movie, especially focusing on the verbs and nouns.\n- Evaluate whether the predicted audio description includes specific details rather than just generic points. It should provide comprehensive information that is tied to specific elements of the video.\n- Consider synonyms or paraphrases as valid matches. Consider pronouns like 'he' or 'she' as valid matches with character names. Consider different character names as valid matches. \n- Provide a single evaluation score that reflects the level of match of the prediction, considering the visual elements like actions, objects and interactions. \n<</SYS>>\n\n{} [/INST] "

# Score adapter: since LLaMA3-8B results in unreasonably low scores (often 0 or 1 out of 5) if the same prompt is used as LLaMA2-7B. We devise to score adapter sentence inserted to LLaMA3-8B prompt that relieves the strictness of scoring. This still leads to consistent scoring.
# For LLaMA2-7B
score_adapter = ""
# For LLaMA3-8B
score_adapter = "Be generous at scoring, and focus on matching of sentence meaning. "

# Insert GT and predicted ADs ("{gt_ad}" and "{pred_ad}"), and the score adapter ("{score_adapter}")
user_prompt = (
    "Please evaluate the following movie audio description pair:\n\n"
    f"Correct Audio Description: {gt_ad}\n"
    f"Predicted Audio Description: {pred_ad}\n\n"
    "Provide your evaluation only as a matching score where the matching score is an integer value between 0 and 5, with 5 indicating the highest level of match. "
    f"{score_adapter}"
    "Please generate the response in the form of a Python dictionary string with keys 'score', where its value is the matching score in INTEGER, not STRING. "
    "DO NOT PROVIDE ANY OTHER OUTPUT TEXT OR EXPLANATION. Only provide the Python dictionary string. "
    "For example, your response should look like this: {'score': }."
)

messages = [
    {"role": "system", "content": sys_prompt},
    {"role": "user", "content": user_prompt},
]
\end{lstlisting}
% \vspace{-0.2cm}
\end{algorithm}